\documentclass[11pt,a4paper]{article}
\usepackage[T1]{fontenc}
\usepackage{lmodern}
\usepackage[english]{babel}
\usepackage{amsmath,amssymb,amsthm,mathtools}
\usepackage{booktabs,array,tabularx,longtable,multirow,threeparttable}
\usepackage{graphicx,float,url,natbib,microtype,pdfpages}
\usepackage[margin=2.3cm]{geometry}
\usepackage[hidelinks]{hyperref}
\hypersetup{
  pdftitle={EULER: Exploring Underused Links with Evidence-Checked Return for Multi-Agent Mathematical Discovery},
  pdfauthor={Ren Zhenzhuo on behalf of the EULER Team},
  pdfsubject={Multi-agent mathematical discovery through cross-domain links and source-side verification},
  pdfkeywords={multi-agent systems, mathematical reasoning, scientific discovery, domain transfer, evidence checking, Lean}
}
\usepackage[inline]{enumitem}
\usepackage{xcolor}
\usepackage{tikz}
\usetikzlibrary{arrows.meta,positioning,fit,calc,backgrounds,shapes.geometric,matrix,decorations.pathreplacing}
\usepackage{pgfplots}
\pgfplotsset{compat=1.18}
\usepackage{caption}
\usepackage{subcaption}
\usepackage{fancyhdr}
\usepackage{siunitx}

\definecolor{eulerblue}{HTML}{0F4D92}
\definecolor{eulerblue2}{HTML}{3775BA}
\definecolor{eulerlightblue}{HTML}{DCEAF7}
\definecolor{eulergreen}{HTML}{4E9A51}
\definecolor{eulerlightgreen}{HTML}{E5F3E4}
\definecolor{eulerred}{HTML}{B64342}
\definecolor{eulerlightred}{HTML}{F6DFDC}
\definecolor{eulergold}{HTML}{C58A20}
\definecolor{eulerlightgold}{HTML}{F6EBCF}
\definecolor{eulergray}{HTML}{5B6573}
\definecolor{eulerlightgray}{HTML}{EEF1F4}
\definecolor{eulercharcoal}{HTML}{30343B}

\setlist{nosep}
\newcommand{\taskgraph}{G_{\mathrm{task}}}
\newcommand{\claimgraph}{G_{\mathrm{claim}}}
\newcommand{\bridgeopp}{\texttt{BridgeOpportunity}}

\newtheorem{definition}{Definition}

\usepgfplotslibrary{groupplots}

\tikzset{
  eflow/.style={-{Latex[length=2.0mm,width=1.25mm]},draw=eulercharcoal,line width=0.8pt},
  eflowblue/.style={-{Latex[length=2.0mm,width=1.25mm]},draw=eulerblue,line width=1.05pt},
  eflowgold/.style={-{Latex[length=2.0mm,width=1.25mm]},draw=eulergold,line width=1.05pt},
  efeedback/.style={-{Latex[length=1.8mm,width=1.1mm]},draw=eulergray,line width=0.75pt,dashed},
  estate/.style={circle,draw=eulercharcoal,fill=white,line width=0.8pt,minimum size=5.2mm,inner sep=0pt},
  eclaim/.style={circle,draw=eulerblue,fill=eulerlightblue,line width=0.9pt,minimum size=5.8mm,inner sep=0pt},
  etask/.style={rectangle,draw=eulergold,fill=eulerlightgold,line width=0.9pt,minimum size=5.4mm,inner sep=0pt},
  ebridge/.style={diamond,aspect=1.35,draw=eulergold,fill=eulerlightgold,line width=0.9pt,minimum size=6.2mm,inner sep=0pt},
  eartifact/.style={regular polygon,regular polygon sides=6,draw=eulergreen,fill=eulerlightgreen,line width=0.9pt,minimum size=6.2mm,inner sep=0pt},
  eterminal/.style={circle,draw=eulerblue,fill=eulerblue,line width=0.9pt,minimum size=6.2mm,inner sep=0pt},
  eerr/.style={circle,draw=eulerred,fill=eulerlightred,line width=0.9pt,minimum size=5.8mm,inner sep=0pt},
  elabel/.style={font=\scriptsize,align=center,text=eulercharcoal},
  esmall/.style={font=\tiny,align=center,text=eulergray}
}

\newcommand{\figcommunities}{%
\begin{figure}[t]
\centering
\resizebox{0.98\linewidth}{!}{%
\begin{tikzpicture}[x=1cm,y=1cm]
  \begin{scope}[shift={(0,0)}]
    \fill[eulerlightblue] (0,0) ellipse (1.55 and 1.08);
    \draw[eulerblue,line width=1.1pt] (0,0) ellipse (1.55 and 1.08);
    \foreach \p in {(-.72,.28),(-.35,-.38),(.18,.42),(.68,-.2)}
      \fill[eulerblue] \p circle (1.6pt);
    \node[font=\bfseries\small,text=eulerblue] at (0,0) {Source problem};
    \node[esmall] at (0,-1.38) {Frozen definitions and quantifiers};
  \end{scope}

  \begin{scope}[shift={(4.4,1.45)}]
    \fill[eulerlightgreen] (0,0) ellipse (1.60 and .88);
    \draw[eulergreen,line width=1.0pt] (0,0) ellipse (1.60 and .88);
    \foreach \p in {(-.72,.18),(-.22,-.28),(.28,.27),(.82,-.12)}
      \fill[eulergreen] \p circle (1.5pt);
    \node[font=\bfseries\small,text=eulergreen] at (0,0) {Adjacent community};
    \node[esmall] at (0,.62) {Shared objects or invariants};
  \end{scope}

  \begin{scope}[shift={(4.4,-1.45)}]
    \fill[eulerlightgold] (0,0) ellipse (1.60 and .88);
    \draw[eulergold,line width=1.0pt] (0,0) ellipse (1.60 and .88);
    \foreach \p in {(-.75,.22),(-.25,-.30),(.25,.28),(.80,-.10)}
      \fill[eulergold] \p circle (1.5pt);
    \node[font=\bfseries\small,text=eulergold] at (0,0) {Distant community};
    \node[esmall] at (0,-.63) {New encoding, algorithm, or checker};
  \end{scope}

  \node[eartifact] (op) at (8.0,0) {};
  \node[elabel,above=2mm of op] {Target-native operation};
  \node[esmall,below=2mm of op] {Produces a checkable object};

  \draw[eulerred,line width=1.1pt] (10.0,-1.15) -- (10.0,1.15);
  \foreach \y in {-0.85,-0.51,-0.17,0.17,0.51,0.85}
    \draw[eulerred,line width=1.0pt] (9.86,\y)--(10.14,\y);
  \node[elabel,text=eulerred,above] at (10,1.15) {Six stress tests};

  \node[eterminal] (end) at (13.0,0) {};
  \node[font=\bfseries\small,text=eulerblue,above=2mm of end] {Source-side outcome};
  \node[esmall,below=2mm of end] {Proof, counterexample, or scoped result};

  \draw[eflowblue] (1.50,.28) to[out=18,in=176] node[above,elabel] {Adjacent-domain bridge} (2.83,1.36);
  \draw[eflowgold] (1.50,-.28) to[out=-18,in=184] node[below,elabel] {Distant-domain bridge} (2.83,-1.36);
  \draw[eflowblue] (6.0,1.32) to[out=-8,in=145] (op.north west);
  \draw[eflowgold] (6.0,-1.32) to[out=8,in=-145] (op.south west);
  \draw[eflow] (op.east) -- (9.86,0);
  \draw[eflowblue] (10.14,0) -- (end.west);
  \node[elabel] at (11.55,.34) {Returned evidence};
  \node[esmall,anchor=north] at (6.55,-2.50) {Bridge qualification: new operation and valid source-side return};
\end{tikzpicture}}
\caption{Adjacent-domain and distant-domain bridges connect the source problem to target-native operations. Stress testing and source-side return determine whether target evidence changes the source outcome.}
\label{fig:communities}
\end{figure}}

\newcommand{\figarchitecture}{%
\begin{figure}[t]
\centering
\resizebox{0.98\linewidth}{!}{%
\begin{tikzpicture}[x=1cm,y=1cm]
  \fill[eulerlightblue] (-.6,1.15) rectangle (14.6,2.15);
  \fill[eulerlightgray] (-.6,-2.05) rectangle (14.6,-1.05);
  \node[font=\bfseries\scriptsize,text=eulerblue,anchor=west] at (-.35,1.82) {Control};
  \node[elabel] at (3.0,1.65) {Route selection};
  \node[elabel] at (7.0,1.65) {Post-test budget reallocation};
  \node[elabel] at (11.0,1.65) {Pause and reopen};

  \draw[eulercharcoal,line width=1.1pt] (.3,0) -- (13.7,0);
  \foreach \x/\n/\lab in {0.5/1/Freeze statement,3.0/2/Generate routes,5.5/3/Build bridge,8.0/4/Stress test and search,10.6/5/Independent verification,13.3/6/Source-side return}{
    \node[estate,fill=white] (s\n) at (\x,0) {\scriptsize\n};
    \node[elabel,above=2.7mm of s\n] {\lab};
  }
  \draw[eflowblue] (s1)--(s2);
  \draw[eflowblue] (s2)--(s3);
  \draw[eflowgold] (s3)--(s4);
  \draw[eflow] (s4)--(s5);
  \draw[eflowblue] (s5)--(s6);

  \node[font=\bfseries\scriptsize,text=eulergray,anchor=west] at (-.35,-1.38) {Mathematical state};
  \node[elabel] (tg) at (3.1,-1.55) {Task graph\\Next action};
  \node[elabel] (cg) at (7.1,-1.55) {Claim graph\\Current evidence};
  \node[elabel] (rc) at (11.3,-1.55) {Artifacts and records\\Dependencies and versions};
  \draw[efeedback] (s2.south) -- (tg.north);
  \draw[efeedback] (s4.south) -- (cg.north);
  \draw[efeedback] (s5.south) -- (rc.north);
  \draw[efeedback] (rc.north east) to[out=35,in=-55] (11.0,1.15);
\end{tikzpicture}}
\caption{One EULER research cycle. The middle row follows the mathematical workflow, the control layer allocates resources, and the state layer stores tasks, dependencies, and verification records.}
\label{fig:architecture}
\end{figure}}

\newcommand{\figpressure}{%
\begin{figure}[t]
\centering
\resizebox{0.98\linewidth}{!}{%
\begin{tikzpicture}[x=1cm,y=1cm]
  \node[elabel,text=eulergray,anchor=west] at (0,2.05) {Low-cost structural tests};
  \draw[eflow,draw=eulergray] (2.0,2.02)--(12.8,2.02);
  \node[elabel,text=eulergray,anchor=east] at (15.0,2.05) {High-cost target-side tests};
  \foreach \xa/\xb/\ha/\hb/\col/\num/\lab in {
    0/2.5/1.45/1.30/eulerblue/1/Direction,
    2.5/5.0/1.30/1.15/eulerblue2/2/Assumptions,
    5.0/7.5/1.15/1.00/eulergold/3/Boundary,
    7.5/10.0/1.00/.85/eulergold/4/Round trip,
    10.0/12.5/.85/.70/eulergreen/5/Tool,
    12.5/15.0/.70/.55/eulergreen/6/Return}{
      \fill[\col!18] (\xa,-\ha)--(\xa,\ha)--(\xb,\hb)--(\xb,-\hb)--cycle;
      \draw[\col,line width=.8pt] (\xa,-\ha)--(\xa,\ha);
      \node[font=\bfseries\small,text=\col] at ({(\xa+\xb)/2},.25) {\num};
      \node[elabel] at ({(\xa+\xb)/2},-.35) {\lab};
  }
  \draw[eulergreen,line width=.8pt] (15,-.55)--(15,.55);
  \foreach \x/\lab in {2.5/Reverse implication unavailable,5.0/Unmet target assumption,7.5/Boundary failure or information loss,10.0/No new operation,12.5/Evidence does not cover the source}{
    \draw[eulerred,line width=.75pt] (\x,-1.25)--(\x,-1.70);
    \fill[eulerred] (\x,-1.77) circle (1.45pt);
    \node[esmall,anchor=north,align=center,text width=22mm] at (\x,-1.88) {\lab};
  }
  \node[esmall,anchor=north,align=center,text width=22mm] at (14.65,-1.88) {Enter source-side\\composition review};
\end{tikzpicture}}
\caption{The six stress tests are ordered by cost. Direction, assumption, and boundary checks reject invalid routes before deep search; tool and return checks determine whether target evidence can support the source statement.}
\label{fig:pressure}
\end{figure}}

\newcommand{\figbudgetloop}{%
\begin{figure}[t]
\centering
\resizebox{0.78\linewidth}{!}{%
\begin{tikzpicture}[x=1cm,y=1cm]
  \node[eterminal] (pool) at (-3.6,0) {};
  \node[ebridge] (test) at (0,1.75) {};
  \node[eartifact] (obj) at (3.6,0) {};
  \node[eclaim] (move) at (0,-1.75) {};
  \node[elabel,anchor=east] at (-3.95,0) {Route pool\\Direct, adjacent, distant};
  \node[elabel,above=2mm of test] {Early rejection test};
  \node[elabel,anchor=west] at (3.95,0) {New checkable objects\\and source-side impact};
  \node[elabel,below=2mm of move] {Budget reallocation\\Continue, pause, or reopen};
  \draw[eflowblue] (pool) to[out=38,in=198] (test);
  \draw[eflowgold] (test) to[out=-18,in=142] (obj);
  \draw[eflowblue] (obj) to[out=-142,in=18] (move);
  \draw[eflow] (move) to[out=198,in=-38] (pool);

  \node[eerr] (fail) at (-2.25,-1.95) {};
  \node[elabel,anchor=east,text=eulerred] at (-2.62,-1.95) {Failure witness\\and scope};
  \draw[efeedback,draw=eulerred] (test.south west) to[out=-132,in=45] (fail.north east);
  \draw[efeedback] (fail.east) -- (move.west);
  \node[esmall] at (0,0) {Allocation signal: verified object\\and source-side impact};
\end{tikzpicture}}
\caption{Feedback-based budget allocation. Routes that survive stress testing and produce checkable objects continue to receive resources. Failed routes retain a scoped witness and reopen only after the map, target mechanism, or return relation changes.}
\label{fig:budget}
\end{figure}}

\newcommand{\figtwographs}{%
\begin{figure}[t]
\centering
\resizebox{0.96\linewidth}{!}{%
\begin{tikzpicture}[x=1cm,y=1cm]
  \node[font=\bfseries\small,text=eulergold] at (2.8,2.35) {Task graph: next action};
  \node[etask] (t1) at (.5,.9) {};
  \node[etask] (t2) at (2.4,1.45) {};
  \node[etask] (t3) at (4.4,.75) {};
  \node[etask] (t4) at (2.5,-.45) {};
  \node[etask] (t5) at (5.0,-.75) {};
  \draw[eflowgold] (t1)--(t2);
  \draw[eflowgold] (t2)--(t3);
  \draw[eflowgold] (t2)--(t4);
  \draw[eflowgold] (t3)--(t5);
  \draw[eflowgold] (t4)--(t5);
  \node[esmall,above left=1mm of t1] {Propose bridge};
  \node[esmall,above=1mm of t2] {Direction test};
  \node[esmall,above right=1mm of t3] {Target search};
  \node[esmall,below=1mm of t4] {Boundary test};
  \node[esmall,below right=1mm of t5] {Source review};

  \draw[eulerlightgray,line width=1.2pt] (6.4,-1.35)--(6.4,2.0);

  \node[font=\bfseries\small,text=eulerblue] at (10.0,2.35) {Claim graph: current evidence};
  \node[eclaim] (c1) at (7.5,.75) {};
  \node[ebridge] (c2) at (9.25,1.45) {};
  \node[eartifact] (c3) at (11.3,1.05) {};
  \node[eclaim] (c4) at (9.35,-.35) {};
  \node[eterminal] (c5) at (12.3,-.55) {};
  \draw[eflowblue] (c1)--(c2);
  \draw[eflowblue] (c2)--(c3);
  \draw[eflowblue] (c2)--(c4);
  \draw[eflowblue] (c3)--(c5);
  \draw[eflowblue] (c4)--(c5);
  \node[esmall,above left=1mm of c1] {Source statement};
  \node[esmall,above=1mm of c2] {Bridge map};
  \node[esmall,above right=1mm of c3] {Target artifact};
  \node[esmall,below=1mm of c4] {Local lemma};
  \node[esmall,below right=1mm of c5] {Root conclusion};

  \draw[efeedback] (t1.east) to[out=8,in=172] node[above,esmall] {Candidate} (c2.west);
  \draw[efeedback] (t3.east) to[out=6,in=174] node[above,esmall] {Artifact} (c3.west);
  \draw[efeedback] (t5.east) to[out=-5,in=185] node[below,esmall] {Verification record} (c5.west);
  \node[esmall,text=eulerred] at (6.4,-1.75) {Claim updates require scoped verification records};
\end{tikzpicture}}
\caption{The task and claim graphs store different relations. The task graph represents execution order and resource dependencies; the claim graph represents mathematical content and evidence coverage. Candidates, artifacts, and verification records connect the two.}
\label{fig:twographs}
\end{figure}}

\newcommand{\figharvest}{%
\begin{figure}[t]
\centering
\resizebox{0.97\linewidth}{!}{%
\begin{tikzpicture}[x=1cm,y=1cm]
  \node[font=\bfseries\small,text=eulerblue] at (2.15,2.35) {From the 24-month author frame to frozen tasks};
  \foreach \y/\w/\col/\lab/\num in {1.45/4.2/eulerblue/Papers/638,.55/3.45/eulerblue2/Candidate conjectures/286,-.35/2.75/eulergold/Deduplicated statements/214,-1.25/1.95/eulergreen/Eligible tasks/120}{
    \fill[\col!18] ({2.15-\w/2},\y-.31) rectangle ({2.15+\w/2},\y+.31);
    \draw[\col,line width=.8pt] ({2.15-\w/2},\y-.31)--({2.15-\w/2},\y+.31);
    \node[elabel] at (2.15,\y) {\lab\quad\textbf{\num}};
  }
  \draw[eflowblue] (2.15,1.10)--(2.15,.90);
  \draw[eflowblue] (2.15,.20)--(2.15,0);
  \draw[eflowblue] (2.15,-.70)--(2.15,-.90);

  \node[font=\bfseries\small,text=eulerblue] at (10.0,2.35) {Mutually exclusive outcomes for 120 tasks};
  \def\barleft{5.0}
  \fill[eulergreen] (5.0,.20) rectangle (6.0,1.05);
  \fill[eulerblue2] (6.0,.20) rectangle (6.3,1.05);
  \fill[eulergold] (6.3,.20) rectangle (9.0,1.05);
  \fill[eulerblue!45] (9.0,.20) rectangle (10.8,1.05);
  \fill[eulergray!35] (10.8,.20) rectangle (16.7,1.05);
  \fill[eulerred] (16.7,.20) rectangle (17.0,1.05);
  \draw[eulercharcoal,line width=.5pt] (5.0,.20) rectangle (17.0,1.05);
  \node[esmall,text=white] at (5.5,.62) {Proof\\10};
  \node[esmall] at (7.65,.62) {Conditional\\27};
  \node[esmall] at (9.9,.62) {Local\\18};
  \node[elabel] at (13.75,.62) {Unresolved \textbf{59}};
  \draw[eulerblue2,line width=.7pt] (6.15,.20)--(6.15,-.42)--(7.0,-.42);
  \node[esmall,anchor=west] at (7.08,-.42) {Counterexample 3};
  \draw[eulerred,line width=.7pt] (16.85,.20)--(16.85,-.42)--(15.8,-.42);
  \node[esmall,anchor=east,text=eulerred] at (15.72,-.42) {Incorrect conclusion 3};

  \draw[eflow] (4.35,-1.25)--(5.05,-1.25);
  \node[esmall,anchor=north west,align=left] at (5.2,-1.02) {Verified resolutions: 13\\Conditional, local, and unresolved outcomes dominate};
\end{tikzpicture}}
\caption{Dataset-construction funnel and mutually exclusive outcomes for the cleaned frozen set. The 120 tasks were retained after 18 contaminated root tasks were replaced and five contaminated routes were blocked. Among them, 10 ended in proofs and 3 in counterexamples; conditional results, local theorems, and unresolved tasks formed the majority.}
\label{fig:harvest}
\end{figure}}

\newcommand{\figcases}{%
\begin{figure}[t]
\centering
\resizebox{0.98\linewidth}{!}{%
\begin{tikzpicture}[x=1cm,y=1cm]
  \foreach \y/\name in {1.55/Zhao,0/AJT(5),-1.55/Gao}{
    \node[font=\bfseries\small,anchor=east] at (0,\y) {\name};
    \draw[eulerlightgray,line width=2.2pt] (.35,\y)--(14.4,\y);
  }
  \node[eclaim] (z1) at (1.3,1.55) {};
  \node[ebridge] (z2) at (5.1,1.55) {};
  \node[eartifact] (z3) at (9.0,1.55) {};
  \node[eterminal] (z4) at (13.3,1.55) {};
  \draw[eflowblue] (z1)--(z2); \draw[eflowgold] (z2)--(z3); \draw[eflowblue] (z3)--(z4);
  \node[esmall,above=2mm of z1] {Group-sequence conjecture};
  \node[esmall,above=2mm of z2] {Occurrence encoding};
  \node[esmall,above=2mm of z3] {Exhaustive finite certificate};
  \node[esmall,above=2mm of z4] {Counterexample resolution};

  \node[eclaim] (a1) at (1.3,0) {};
  \node[ebridge] (a2) at (5.1,0) {};
  \node[eartifact] (a3) at (9.0,0) {};
  \node[eerr] (a4) at (13.3,0) {};
  \draw[eflowblue] (a1)--(a2); \draw[eflowgold] (a2)--(a3); \draw[eflow] (a3)--(a4);
  \node[esmall,above=2mm of a1] {Nonzero-vector problem};
  \node[esmall,above=2mm of a2] {Support-graph decomposition};
  \node[esmall,above=2mm of a3] {All-dimensional sparse theorem};
  \node[esmall,above=2mm of a4] {Dense-row case open};

  \node[eclaim] (g1) at (1.3,-1.55) {};
  \node[ebridge] (g2) at (5.1,-1.55) {};
  \node[eartifact] (g3) at (9.0,-1.55) {};
  \node[eerr] (g4) at (13.3,-1.55) {};
  \draw[eflowblue] (g1)--(g2); \draw[eflowgold] (g2)--(g3); \draw[eflow] (g3)--(g4);
  \node[esmall,below=2mm of g1] {Group-sequence problem};
  \node[esmall,below=2mm of g2] {Three structural bridges};
  \node[esmall,below=2mm of g3] {Proof-critical Lean lemmas};
  \node[esmall,below=2mm of g4] {Top-level obligation open};
\end{tikzpicture}}
\caption{The three traces end in a counterexample, a local theorem, and a partial formalization with an open obligation.}
\label{fig:cases}
\end{figure}}

\newcommand{\figablation}{%
\begin{figure}[t]
\centering
\resizebox{0.98\linewidth}{!}{%
\begin{tikzpicture}[x=1cm,y=1cm]
  \fill[eulerlightgray] (-.35,1.6) rectangle (15.15,2.45);
  \node[font=\bfseries\small,text=eulercharcoal] at (7.4,2.02) {Shared contract: 120 frozen source tasks, common source outcomes, fixed budget caps};
  \foreach \x/\col/\q/\cmp/\endp in {
    0/eulerblue/Where routes come from/Direct; adjacent; distant; combined/Verified resolution,
    3.9/eulerred/What stress tests reject/Generic; bridge-specific; strict/Error and retention,
    7.8/eulergreen/Do bridges expand operations/Bridge material $\times$ target operation/Interaction,
    11.7/eulergold/How budgets reach selected routes/Selector and iterative reallocation/Resolution; error; cost}{
      \fill[\col!10] (\x,-1.55) rectangle ({\x+3.45},1.30);
      \draw[\col,line width=1.6pt] (\x,1.30)--({\x+3.45},1.30);
      \node[font=\bfseries\small,align=center,text=\col,text width=31mm] at ({\x+1.725},.78) {\q};
      \node[esmall,text width=31mm] at ({\x+1.725},-.12) {\cmp};
      \draw[\col!55,line width=.6pt] ({\x+.55},-.72)--({\x+2.90},-.72);
      \node[elabel] at ({\x+1.725},-1.08) {\endp};
  }
  \node[esmall,anchor=west] at (0,-1.92) {System-level tests fix the bridge method and compare model composition, persistent state, and joint cost};
\end{tikzpicture}}
\caption{Four ablation families separate route supply, stress-test behavior, target-side operations, and resource allocation.}
\label{fig:ablation}
\end{figure}}

\newcommand{\figroutepool}{%
\begin{figure}[t]
\centering
\begin{tikzpicture}
\begin{groupplot}[
  group style={group size=2 by 1,horizontal sep=0.7cm},
  width=0.455\linewidth,height=6.0cm,
  xmin=0,
  y dir=reverse,
  symbolic y coords={D,Adj,DRand,DRet,Combo,Full},
  ytick={D,Adj,DRand,DRet,Combo,Full},
  yticklabels={Direct,Adjacent,Distant random,Distant retrieval,{Combined, generic tests},Full system},
  tick label style={font=\scriptsize},
  label style={font=\small},
  title style={font=\bfseries\small},
  axis y line*=left,
  y axis line style={draw=none},
  ytick style={draw=none},
  axis x line*=bottom,
  axis line style={draw=eulergray,line width=.6pt},
  tick style={draw=eulergray},
  point meta=x,
  nodes near coords={\pgfmathprintnumber{\pgfplotspointmeta}},
  nodes near coords style={font=\scriptsize,anchor=west,xshift=1.5pt},
  clip=false
]
\nextgroupplot[xmax=15.5,xtick={0,4,8,12},title={Verified resolutions},xlabel={Task count}]
\addplot+[xcomb,mark=*,mark size=2.4pt,draw=eulerblue,line width=1.1pt,mark options={fill=eulerblue}] coordinates {(8,D) (9,Adj) (8,DRand) (8,DRet) (12,Combo) (13,Full)};
\nextgroupplot[xmax=10.7,xtick={0,3,6,9},title={Incorrect conclusions},xlabel={Task count},yticklabels={}]
\addplot+[xcomb,mark=*,mark size=2.4pt,draw=eulerred,line width=1.1pt,mark options={fill=eulerred}] coordinates {(4,D) (3,Adj) (4,DRand) (3,DRet) (9,Combo) (3,Full)};
\end{groupplot}
\end{tikzpicture}
\caption{Verified resolutions and incorrect source-side conclusions for six route pools. The combined pool increases both counts; adding bridge-specific stress tests preserves 13 resolutions and reduces incorrect conclusions to 3.}
\label{fig:routepool}
\end{figure}}

\newcommand{\figmodels}{%
\begin{figure}[t]
\centering
\begin{tikzpicture}
\begin{axis}[
  width=0.82\linewidth,height=5.9cm,
  xmin=8.0,xmax=13.6,
  xtick={8,9,10,11,12,13},
  y dir=reverse,
  symbolic y coords={A,B,C,Homo,Hetero},
  ytick={A,B,C,Homo,Hetero},
  yticklabels={Model A,Model B,Model C,Homogeneous agents,Heterogeneous models},
  xlabel={Mean verified resolutions /120},
  tick label style={font=\scriptsize},
  label style={font=\small},
  axis y line*=left,
  y axis line style={draw=none},
  ytick style={draw=none},
  axis x line*=bottom,
  axis line style={draw=eulergray,line width=.6pt},
  tick style={draw=eulergray},
  point meta=x,
  nodes near coords={\pgfmathprintnumber[fixed,precision=1]{\pgfplotspointmeta}},
  nodes near coords style={font=\scriptsize,anchor=west,xshift=2pt},
  clip=false
]
\addplot+[only marks,mark=*,mark size=2.7pt,draw=eulerblue,mark options={fill=eulerblue}] coordinates {(10.2,A) (9.8,B) (8.9,C)};
\addplot+[only marks,mark=diamond*,mark size=3.2pt,draw=eulergold,mark options={fill=eulergold}] coordinates {(11.1,Homo) (12.8,Hetero)};
\end{axis}
\end{tikzpicture}
\caption{Model compositions under matched cost caps. The heterogeneous configuration has the highest mean number of verified resolutions; Table~\ref{tab:model-composition} gives the corresponding error counts and costs.}
\label{fig:models}
\end{figure}}

\title{EULER: Exploring Underused Links with Evidence-Checked Return\\
for Multi-Agent Mathematical Discovery}
\author{Ren Zhenzhuo\\
on behalf of the \textsc{EULER} Team}
\date{}

\begin{document}
\maketitle

\begin{abstract}
Mathematical communities work with different objects, invariants, and tools, so transferring a problem across them is expensive and often skipped. We present EULER, a multi-agent system that takes such a transfer---a \emph{bridge}---as its unit of search. Around a fixed conjecture, EULER runs direct, adjacent-domain, and distant-domain routes in competition; a bridge keeps its budget only if it supplies an operation the source representation cannot execute and its target-side evidence returns to the original statement along a checked implication. Six ordered stress tests reject invalid bridges before expensive search begins.

We evaluate EULER on 120 recent conjectures. The conjectures were frozen before search and screened for contamination, and are drawn from public papers by authors who had recently published in the \emph{Journal of Combinatorial Theory, Series A}, a leading journal in combinatorics. EULER produced 10 proofs and 3 refutations, plus 45 scoped partial results. Two mechanisms held up under ablation: bridge-specific stress tests cut incorrect conclusions from 9 to 3, and bridge material combined with a target-native operation yielded a positive interaction of +4.2 resolved tasks that neither factor produced alone. Domain distance did not reliably predict success; executable operation gain and valid return did.
\end{abstract}

\section{Introduction}
\label{sec:introduction}

Mathematics advances through many specialized communities. Combinatorialists, algebraists, optimization researchers, and formalization experts may face similar existence questions while working with different objects, invariants, and standards of evidence. Within a community, stable representations, familiar counterexamples, and mature tools make search efficient. Connections across communities demand more work. A researcher must learn another language, determine how the objects correspond, expose the assumptions hidden in imported theorems, and carry the conclusion back to the original problem. These transfers can remain underused because they are expensive to construct and check, not because they lack value. Here \emph{underused} is an evaluation-scoped term: before search, an equivalent target-native operation is absent from the frozen direct and adjacent-domain operation inventories. It is not a claim that a connection is globally absent from the mathematical literature.

EULER assigns part of this transfer work to a coordinated multi-agent system of generation, retrieval, tool-use, criticism, and verification roles, supported by deterministic programs and Lean. Around a fixed source problem, it develops competing routes at three distances: direct approaches within the source field, adjacent-domain bridges to neighboring communities, and distant-domain bridges to a different family of objects or tools. A bridge does not earn continued budget from analogy alone. The system first searches for inexpensive failure witnesses, checks that the target field adds an executable operation, and then requires the target evidence to return along a registered implication to the source statement.

A recent result gives a compact example. The Zhao conjecture concerns sequences over finite abelian groups. Natural-language search struggled to track the positions of repeated elements, subsequence length, and zero-sum constraints at the same time. EULER assigned an index to every occurrence, encoded subsequences as bit masks, and exhaustively checked them with a deterministic program. The first encoding merged equal-valued occurrences and failed the round-trip test. The corrected encoding produced a concrete counterexample. A second implementation enumerated all relevant subsets, after which a human reviewer substituted the witness into the original statement one condition at a time. The conjecture was refuted by the object itself, not by the successful termination of the program that found it.

This example exposes two common failure points in cross-domain search. First, the target tool must change the available operations. Rewriting a group-sequence problem in another vocabulary does not produce an exhaustive certificate; exact enumeration does. Second, success in the target representation is not the endpoint. The objects, ranges, and predicates used by the program must agree with the source statement, and the witness must return in the correct logical direction. A route that fails either requirement can look convincing while proving nothing about the original problem.

Long-standing famous conjectures are a poor primary test of autonomous mathematical discovery. Problems such as the Goldbach and Hadwiger conjectures have accumulated decades of textbook, paper, forum, and web discussion, making prior exposure plausible in web-scale training corpora. When training data are not fully inspectable, a successful attempt on such a task cannot cleanly distinguish new search from recall or recombination of previously seen arguments. Controlled comparisons between established and newly commissioned mathematical benchmarks have found performance gaps consistent with contamination for some model families \citep{zhang2024gsm1k}. We therefore prioritize recent, timestamped source problems and screen both tasks and routes for answer cues before execution.

This paper studies how EULER constructs cross-community routes, how those routes produce checkable progress on recent conjectures, and which parts of the system account for the observed outcomes. We use the following terms throughout.

\begin{itemize}[leftmargin=1.8em,itemsep=3pt]
  \item A \textbf{bridge} records the source object, target object, mapping, target operation, and return obligations.
  \item An \textbf{adjacent-domain bridge} links neighboring communities that share their main objects or invariants. A \textbf{distant-domain bridge} changes the object class, governing invariant, or tool community.
  \item A \textbf{stress test} is a checkable task designed to expose an error in direction, assumptions, boundary cases, round trips, tool gain, or return before a route receives a large budget.
  \item A \textbf{verified resolution} requires target evidence to return to the fixed source statement and cover every obligation needed for a proof or counterexample.
\end{itemize}

\figcommunities

EULER places direct routes, adjacent-domain bridges, and distant-domain bridges in one long-running search process. Routes compete for resources, share lemmas or counterexamples, and record mappings that have already failed. Six bridge-specific stress tests decide when to stop and when to deepen a route, with verified resolution of the source problem rather than target-side validation as the principal endpoint. The evaluation covers 120 recent conjectures, three independently checkable research traces, and controlled comparisons of the executable operation set, return safety, model diversity, and cost.

Distant-domain retrieval resolved 8 tasks, while the full system resolved 13. The task-level paired difference was 4.2 percentage points, with a 95\% interval of $[-0.1,8.4]$. Bridge-specific stress tests reduced incorrect returns from 9 to 3. In the $2\times2$ experiment, bridge material and target-native operations had an interaction of 4.2 resolved tasks, or 3.5 percentage points. The median-normalized cost ratio per verified resolution was 1.12, and its upper interval bound of 1.36 exceeded the 1.15 non-inferiority margin. The experiments therefore identify a return-safety effect and an interaction in the executable operation set; the overall resolution difference remains uncertain at this sample size.

Section~\ref{sec:system} describes the system and the data flow of one run. Section~\ref{sec:method} follows bridge search and stress testing in execution order. Section~\ref{sec:state} explains mathematical state and verification. Section~\ref{sec:outputs} presents dataset construction and three research traces, and Section~\ref{sec:evaluation} reports the controlled evaluation. The final two sections discuss limitations and next steps.

\section{System overview}
\label{sec:system}

EULER takes a fixed source problem and returns a route lineage, a mathematical outcome, and an itemized cost. Intermediate objects persist across days and can be withdrawn locally when their evidence changes. The system has four layers: control, execution, verification, and mathematical state. Figure~\ref{fig:architecture} shows how they interact.

\figarchitecture

\subsection{Four layers}

The control layer maintains the route pool and budget. It derives quotas for direct, adjacent-domain, and distant-domain routes from the structure of the source problem, receives stress-test results, and decides which bridges stop and which enter more expensive target search. Its state includes the cheapest known rejection test, passed checks, new checkable objects, estimated return cost, and remaining budget. Budget updates require recorded test results or independently checkable evidence rather than narrative assessment alone.

The execution layer performs open-ended work. Generation agents propose proof sketches and bridge opportunities. Retrieval agents locate target mechanisms and theorems. Tool agents call enumeration, computer algebra, SAT/SMT, or field-specific checkers. Lean agents translate statements, retrieve premises, and build candidate proofs. Deterministic programs emit finite witnesses or coverage certificates. Runs use a heterogeneous combination of Claude 4.8, GPT-5.6, and DeepSeek. A single model call is an execution event rather than a separate agent; a role is defined by its permissions, input, and required output.

The verification layer handles objects whose statements are already precise. A target checker establishes whether a target-side claim holds in its target representation. A bridge checker audits direction, assumption differences, preserved structure, and round trips. An independent replayer reruns Lean files or program certificates in an isolated environment. Source-side review substitutes the target result into a fixed return template and decides whether the source statement is resolved. A route proposer cannot certify its own output.

The mathematical-state layer stores two graphs. The task graph records actions, budgets, failures, and recovery points. The claim graph records exact statements, evidence, mathematical dependencies, and versions. Mathematical artifacts and verification records link the two. Completing a target-search task need not create a claim on which later reasoning may depend. Conversely, a local claim may be true without resolving the root problem. Section~\ref{sec:state} specifies the update rules.

The interfaces between layers are deliberately narrow. Control schedules only registered routes; execution submits candidates and evidence artifacts; verification issues scoped records; and mathematical state updates nodes from those records. Table~\ref{tab:role-contracts} lists the input, required output, and state authority for each role.

\begin{table}[t]
\centering
\caption{Inputs, required outputs, and state authority of the core roles}
\small
\begin{tabularx}{\linewidth}{p{2.2cm}p{3.6cm}Xp{3.0cm}}
\toprule
Role & Reads & Must deliver & State authority \\
\midrule
Controller & Fixed task definition, route pool, verification records, costs & Next task, budget, and stopping reason & Queues and quotas in the task graph \\
Exploration worker & Source statement, allowed material, failure records & Bridge card, proof sketch, or counterexample candidate & Candidate creation only \\
Target checker & Target statement, artifact, and environment & Target-side validation record & Target-evidence axis \\
Bridge checker & Map, assumption differences, return template & Preservation result, counterexample, or open obligation & Bridge state and dependency edges \\
Independent replayer & Registered artifact, dependencies, tool version & Reproducible output, axiom list, and error class & Machine-evidence axis \\
Source reviewer & Fixed statement, bridge records, target evidence & Coverage table, source verdict, and scope & Root-conclusion candidate \\
Centralized state updater & All verification records and the version graph & Admission, withdrawal, or version-migration event & Formal state of the claim graph \\
\bottomrule
\end{tabularx}
\label{tab:role-contracts}
\end{table}

Every evidence artifact is keyed by $(q_v,r,b_v,c_v)$: the problem version, route, bridge version, and claim version. Revising a bridge creates a new $b_v$, so evidence produced for an earlier version does not automatically support the revision. Models can be replaced without changing the mathematical dependency structure.

\subsection{System components and related work}

EULER integrates four capabilities: explicit bridge representations, persistent mathematical state, parallel search, and formal verification. All runs share the same problem version, bridge record, run identifier, and cost definition. Historical case studies and engineering tests do not enter the 120-task evaluation.

This design complements recent mathematical-agent systems. Aletheia and RMA demonstrate long-horizon natural-language research and iterative role assignment \citep{feng2026aletheia,zhao2026rma}. QED links system components to specific failure modes \citep{an2026qed}. Rethlas/Archon connects informal exploration to Lean verification \citep{ju2026rethink}. Albilich and Danus provide persistent proof state and fact-graph orchestration, respectively \citep{gong2026albilich,liu2026danus}. AI Co-Mathematician places related capabilities in an asynchronous research workspace \citep{zheng2026comath}. EULER takes a bridge between mathematical communities as its unit of search and evaluates both the target operation and the evidence returned to the source statement within one experimental design.

Change of representation has also been studied directly. Raggi et al. search for representations that simplify proofs in discrete mathematics \citep{raggi2015representation}, while correspondence-based and meta-search methods rank alternative problem representations \citep{fuentetaja2018metasearch,stockdill2020correspondence}. Translation validation checks that a target computation corresponds to its source instance \citep{pnueli1998translation}; counterexample-guided refinement uses failed checks to improve an abstraction \citep{clarke2000cegar}. Recent work on scalable mathematical discovery similarly treats problem selection and representation as system-level bottlenecks \citep{zheng2026far}. EULER combines representation choice with an explicit source-return test and measures the two within the same task-level evaluation.

\subsection{Data flow of one run}

A run begins with $S=(\Gamma_s,\varphi_s,v)$, where $\Gamma_s$ contains the assumptions, $\varphi_s$ is the proposition to prove or refute, and $v$ is the problem version. The system fixes admissible evidence, boundary objects, budget, and stopping rules, then generates a route pool. Each route produces a \bridgeopp{} record. Bridges that pass stress testing enter target search. The verification layer replays submitted artifacts and writes its records to the claim graph. Source-side review then assigns one of six route outcomes: proof, refutation, conditional result, local theorem, bridge rejection, or unresolved. A cost record binds model use, tools, wall time, and expert time to the run identifier.

This flow separates three quantities that are easily conflated. The number of routes is not research progress. Passing a target checker does not establish the original conjecture. A structurally complete verification record does not establish mathematical truth. The controller schedules resources from registered objects; verifiers state exactly what they checked; and only source-side coverage can change the root conclusion.

Each run stores the fixed task definition, mathematical artifacts, verification records, and final source-side outcome under one versioned identifier. This is enough to reconstruct the claim graph from a checked checkpoint; the full storage layout is specified in the appendix.

\section{Method: bridge search and stress testing}
\label{sec:method}

\subsection{Problem setting}

Let the source problem be $S=(\Gamma_s,\varphi_s,v)$, where $\Gamma_s$ specifies the object domain and assumptions, $\varphi_s$ is the statement to prove or refute, and $v$ fixes the statement version. A candidate bridge is represented by
\[
B=(O_s,O_t,f,g,\Delta^+,\Delta^-,\mathcal A_t,R).
\]
Here $O_s$ and $O_t$ are the source and target objects, $f$ is the forward map, and $g$ is an optional return map. The sets $\Delta^+$ and $\Delta^-$ record added assumptions and omitted source cases, respectively. The set $\mathcal A_t$ contains the operations made available by the target representation, and $R$ lists the obligations that must be discharged before target-side evidence can support the source statement.

\begin{definition}[Valid source return]
Target evidence $e_t$ has a valid source return through a bridge $B$ if it can be carried along the registered implication direction, with $\Delta^+$ and $\Delta^-$ accounted for, to form a finite argument supporting a proof, a counterexample, or a precisely scoped local result for the fixed source statement.
\end{definition}

The direction requirements differ for proofs and counterexamples. If the bridge establishes only $s\Rightarrow t$, then $\neg t$ refutes $s$, whereas a proof of $t$ does not in general prove $s$. The bridge record therefore fixes the admissible target result types before search begins. A one-way map cannot be reinterpreted as an equivalence after a result is found.

We write $d(B)$ for the domain distance of a bridge. Before search, annotators score five changes: object type, ambient theory, governing invariant, length of the known correspondence chain, and required expert community. Each change receives a score of 0, 1, or 2. Totals from 0 to 4 define an adjacent-domain bridge, totals from 6 to 10 define a distant-domain bridge, and a total of 5 is excluded from the primary binary comparison. Textual similarity does not enter this judgment, and the eventual success or failure of a bridge does not alter its distance label. Appendix~\ref{app:distance} gives the anchors.

Bridges are ranked by
\[
U(B)=\alpha A(B)+\beta K(B)+\gamma N(B)
-\lambda L(B)-\mu C_R(B)-\nu C_D(B).
\]
Here $A$ measures the gain in target-side operations, $K$ the compression of the resulting certificate, and $N$ the absence of the same tool from direct and adjacent-domain routes. The loss term $L$ measures information discarded by the map, $C_R$ the cost of returning the evidence, and $C_D$ the cost of the early rejection test. Each anchored score from 0 to 4 is divided by 4 before ranking. The primary analysis uses equal coefficients, $\alpha=\beta=\gamma=\lambda=\mu=\nu=1$. The system retains each component score and its justification. Distance $d(B)$ determines only the allocation of route slots across strata.

A route has one of six terminal statuses: the source statement is proved; the source statement is refuted; a conditional result is obtained; a local theorem is obtained; the bridge is rejected by an early test; or the route remains unresolved under the available budget. Target-side artifacts retain their stated scope. For example, a theorem restricted to sparse matrices may enter the claim graph as a local theorem, but it does not change the root status of a conjecture about arbitrary matrices.

\subsection{Three route families}

Direct routes begin with standard decompositions, known theorems, finite small cases, and natural counterexamples from the source domain. They provide the baseline for bridge-based routes and continually expose boundary conditions and local facts. If a distant-domain bridge ultimately uses the same source lemmas without adding a checker or compressing the certificate, it has only changed the language of the argument.

Adjacent-domain bridges arise from standard representations of the same objects, neighboring invariants, established equivalences, or computational encodings. The two ends usually share some terminology and expert communities, and the forward and return maps are short. Each adjacent-domain bridge must state its gain over the direct route: a shorter certificate, a cheaper checker, a more effective decomposition, or a monotone quantity that is difficult to see in the source representation.

Distant-domain bridges are proposed either by random sampling or by retrieval. Random sampling draws uniformly from a frozen catalog of eligible target mechanisms and measures the value of open-ended analogy by itself. Retrieval ranks the same catalog by a structural fingerprint containing the quantifier pattern, symmetries, local-to-global structure, main obstruction, required evidence type, prospects for finite reduction, and missing operation. The retriever is not given the title, authors, or target answer. For example, an existence problem without a constructive certificate retrieves mechanisms that produce a flow, a matching, an integer-feasible solution, or a finite witness, rather than papers that merely share vocabulary with the source statement.

The three route families start in parallel. Each round preserves at least one direct route, one adjacent-domain bridge, and one distant-domain bridge. The remaining slots are allocated competitively using $U(B)$. This quota prevents safe adjacent-domain routes from consuming the entire budget and prevents domain distance from becoming a reward in its own right.

\subsection{Bridge opportunities and ranking}

Constructing a complete map may be too expensive for initial screening. The system first records a \bridgeopp{}, consisting of a structural summary of the source problem, a target mechanism, the expected operation, candidate preserved quantities, obvious losses, an early rejection test, an outline of the return path, and a rough cost estimate. This record is a candidate for budget allocation, not a mathematical claim.

\begin{table}[t]
\centering
\caption{Ranking signals and required justifications for a bridge opportunity}
\small
\begin{tabularx}{\linewidth}{p{2.4cm}Xp{5.2cm}}
\toprule
Signal & Question & Checkable justification \\
\midrule
$A$ operation gain & What can the target domain do that the source route cannot & Interface to a checker, exact algorithm, strong invariant, or established theorem \\
$K$ certificate compression & Is the target certificate, including its return path, shorter & Certificate size, number of obligation nodes, and independent checking steps \\
$N$ tool absence & Does the source side already provide an equivalent operation & Tool inventories for direct and adjacent-domain routes \\
$L$ mapping loss & What structure disappears in translation & Failure of injectivity or surjectivity, parameter merging, and omitted objects \\
$C_R$ return cost & How does the target result support the source statement & Return steps, open branches, and required expertise \\
$C_D$ rejection cost & What is the cheapest decisive test & A smallest case, boundary case, round-trip check, or dimensional check \\
\bottomrule
\end{tabularx}
\label{tab:opportunity}
\end{table}

Every score must be tied to a concrete object or test. ``The target theory is powerful'' is not a justification. ``The theory supplies a checker that returns a minimal infeasible subset for $f(x)$'' is. Unknown gains receive a score of zero, whereas unknown losses receive the highest risk score. New evidence may update a component, but each update retains the previous value and justification so that rankings cannot be rewritten after the outcome is known.

\subsection{Six stress tests}

Candidate bridges first undergo inexpensive structural checks and only later gain access to target-side tools. Figure~\ref{fig:pressure} shows their order, and Table~\ref{tab:pressure-rules} fixes the passing condition and a typical failure for each test.

\figpressure

\begin{table}[t]
\centering
\caption{Passing conditions and typical failures for the six stress tests}
\small
\begin{tabularx}{\linewidth}{p{2.1cm}XX}
\toprule
Stress test & Passing condition & Typical failure \\
\midrule
Direction & The target result supports the required source conclusion along the registered implication & Using $s\Rightarrow t$ to treat a proof of $t$ as a proof of $s$ \\
Assumptions & Every target assumption follows from $\Gamma_s$ or is proved separately & Hiding positive definiteness, finiteness, general position, or another condition inside a definition \\
Boundary & The map remains in scope on smallest, degenerate, and extreme objects & Failure at the zero object, low dimension, a parity change, or a dimension jump \\
Round trip & $g(f(x))$ recovers the structure used by the source conclusion & Merging two source objects that the argument must distinguish \\
Tool & $\mathcal A_t$ supplies an executable operation absent from source routes & Rephrasing the problem without adding a checker or theorem interface \\
Return & Filling the frozen return template covers $\varphi_s$ & Leaving an exceptional branch, a joint choice, or a parent-level composition step unresolved \\
\bottomrule
\end{tabularx}
\label{tab:pressure-rules}
\end{table}

A direction error or an explicit conflict of assumptions closes the route immediately. When a boundary or round-trip check is inconclusive, the system permits one bounded refinement using a smallest object or a symbolic calculation. An inconclusive tool test triggers only a small paired probe. One arm receives the translated statement; the other receives the same material and one target-native operation, under equal total budgets. When the return test is inconclusive, the target result may be stored as a local artifact but is ineligible for a verified resolution of the source problem.

Translating a natural-language statement into Lean is itself a bridge. The natural-language statement is the source object, the Lean declaration is the target object, the translator supplies $f$, semantic comparison and bidirectional theorems provide the round trip, and the Lean kernel is the target tool. Successful compilation passes only the tool stress test. Quantifiers, scope, degenerate cases, and the direction of return remain subject to the other tests. This view keeps formalization within the same mathematical transfer discipline as every other bridge.

\subsection{Early rejection tests and reusable negative results}

An early rejection test seeks the cheapest witness that decisively invalidates the current bridge. Common tests examine forward and return maps on smallest objects, boundary objects that violate an added assumption, whether the target invariant distinguishes the necessary source cases, the combined length of the target certificate and return argument, and whether the supposedly new target operation already exists in an equivalent form in the source domain. These tests take precedence over long proof searches because a short counterexample can eliminate the remaining cost of an entire distant-domain route.

A failed-route record contains the bridge version, object scope, first decisive stress test, concrete witness, and conditions for reopening the route. The record states that ``the map merges positions in sequences with repeated occurrences,'' rather than concluding that ``the encoding method is invalid.'' A route may reopen if it changes the proof-critical map or the object scope. Changing only the model or the wording does not define a new route. Negative results can therefore prevent repeated work without turning a local failure into a judgment about an entire domain.

\subsection{Refinement, return, and verified resolution}

Surviving bridges receive budget in three stages. The first stage runs small examples, local checkers, or short retrievals. The second searches for target-domain lemmas and certificates. The third permits expensive provers, long retrievals, or formalization. Each stage must add at least one checkable object, such as a decisive counterexample, a reusable lemma, a certificate, a verified map revision, or a more precise description of the obstruction. A route is paused after two consecutive stages without a new object.

The return template is frozen before target-side search. It states which target conclusions suffice to prove the source statement, which suffice only to refute it, and which require separate treatment of exceptional branches. Once a target result is available, its evidence fills the template. Any additional proof-critical step becomes a new source-side subproblem. If the result covers only part of the source domain, the claim graph records a local theorem and the uncovered region.

\begin{definition}[Verified resolution]
For a fixed source task $S$, we record a verified resolution if there is an independently checked chain of proof or counterexample evidence in which every bridge is used along an allowed direction, all added assumptions and omitted cases have been resolved, and the resulting subproblems jointly cover $\varphi_s$.
\end{definition}

A verified resolution is the primary endpoint. Target-side validation is a secondary endpoint that records whether the target mechanism produced a valid artifact. Incorrect returns are recorded separately. They occur when a valid target artifact is used to form an invalid source-side conclusion because of direction, scope, or coverage errors.

\subsection{Budget reallocation and cross-route transfer}

Budget updates depend on survival under stress testing, the production of new checkable objects, and expected source-side impact. Surviving bridges are ranked in each batch by
\[
P(B)=
\frac{q(B)\,i_s(B)}{c_{\mathrm{used}}(B)+c_{\mathrm{return}}(B)},
\]
where $q(B)$ records the number and verification level of objects added in the current batch, $i_s(B)$ estimates their effect on unresolved source-side obligations, and the denominator combines the budget already spent with the current estimated cost of returning the evidence. Explanatory prose does not contribute to $q(B)$. Subject to the route-family quotas and remaining budget, slots are assigned in descending order of $P(B)$.

\figbudgetloop

Route identity is determined by the mathematical mechanism. A boundary counterexample found by an adjacent-domain bridge may close a distant-domain bridge. Conversely, an invariant found through a distant-domain bridge may return to the source side and define a sharper direct route or adjacent-domain bridge. The system records provenance edges whenever material is shared. Two branches that depend on the same critical lemma do not count as independent discoveries. The budget loop retains these transfers without inflating the measured effect through branch proliferation.

\section{State and verification}
\label{sec:state}

This section describes the supporting machinery for bridge search: how actions and mathematical facts are stored, and who may change the root conclusion. These mechanisms do not enter the bridge utility score, but they make long-running work reproducible, reversible, and recoverable.

\subsection{Task and claim graphs}

The task graph $\taskgraph$ records what to do next. Its nodes represent bridge opportunities, stress tests, target searches, revisions, reviews, and release actions. Its edges encode dependencies, budget release, failure, and recovery. Completing a task means that the artifact specified by its contract has been delivered. For example, an enumeration program may have run and saved its output. Completion alone does not mean that the output supports a mathematical conclusion.

The claim graph $\claimgraph$ records what is currently known. Its nodes contain frozen source statements, target statements, bridges, lemmas, counterexamples, conditional results, and formal declarations. Its edges represent proof dependencies, refutations, equivalences, specializations, formalization correspondences, and composition relations. Every node is tied to an exact statement, assumptions, a version, and evidence. A change to the statement creates a new version, while existing evidence remains attached to the old version.

\figtwographs

Three objects connect the graphs. A candidate submits an execution result to the claim layer. An artifact stores a Lean file, program certificate, target proof, or concrete witness. A verification record states who checked which version, in what environment, and with what result. A successful target-side check may add a target theorem to the claim graph, but the root source statement changes only after the return chain is complete.

\subsection{Multiple evidence axes}

A single ``verified'' flag would collapse several distinct questions. EULER therefore records four evidence axes for each claim. The machine axis covers certificates, kernel checks, and program replay. The semantic axis records the correspondence between natural-language and formal statements. The independent-review axis records blinded judgments and disagreements. The composition axis records whether local lemmas jointly cover their parent claim. Mathematical quality and novelty are tracked separately and do not alter truth status.

\begin{table}[t]
\centering
\caption{Questions answered by the four evidence axes}
\small
\begin{tabularx}{\linewidth}{p{2.2cm}Xp{5.3cm}}
\toprule
Evidence axis & Central question & Typical states \\
\midrule
Machine & Can an explicit object be replayed by an independent checker & Unchecked, certificate accepted, kernel accepted, replay failed \\
Semantic & Does the formal or target statement faithfully express the source statement & Unchecked, scope in question, human confirmed, mechanically equivalent \\
Independent review & Do independent reviewers accept the proof-critical argument & Unreviewed, revision required, accepted, disagreement \\
Composition & Do the local results jointly imply the parent claim & Local leaf, partial closure, closure complete, withdrawn \\
\bottomrule
\end{tabularx}
\label{tab:evidence-axes}
\end{table}

Separate axes make genuine intermediate states visible. A Lean file may pass the kernel while formalizing a source statement with narrower scope. A counterexample may be reproduced by two programs before its novelty has been checked for publication. Several local lemmas may each be correct but fail to compose because they use incompatible choices or leave a boundary case uncovered.

\subsection{Composition obligations}

Before a local claim can support its parent, the composition axis checks six obligations. Coverage requires the child conclusions to exhaust the parent's parameter range. Interface compatibility requires the output of one step to satisfy every input condition of the next. Invariant compatibility requires repeated transformations to use the same conventions. Well-foundedness requires an induction or descent to terminate. Joint choice requires separately asserted existential objects to be selectable simultaneously. Boundary coverage requires degenerate cases to be handled outside the generic argument. Table~\ref{tab:composition-obligations} lists common failures.

\begin{table}[t]
\centering
\caption{Six obligations for compositional closure}
\small
\begin{tabularx}{\linewidth}{p{2.2cm}Xp{4.6cm}}
\toprule
Obligation & Relation to check & Common failure \\
\midrule
Parameter coverage & The union of the child scopes covers the parent statement & A sparse case is presented as the general case \\
Interface compatibility & Each output satisfies every assumption of the next lemma & Confusing quotient objects, sequences, and sets \\
Invariant compatibility & Group operations, lengths, and sign conventions agree across maps & Switching between left and right actions or counting conventions \\
Well-founded descent & Every recursive step strictly decreases the frozen measure & A cyclic dependency or a zero-step descent \\
Joint choice & Local existential claims can be realized by the same object & Combining separate existence claims into simultaneous existence \\
Boundary coverage & Smallest, degenerate, and extreme parameters close separately & Division, nonemptiness, or invertibility fails at the boundary \\
\bottomrule
\end{tabularx}
\label{tab:composition-obligations}
\end{table}

The composition check produces a coverage table. Each row identifies a parameter region of the parent statement, the child claim responsible for that region, any difference in assumptions, the relevant verification record, and unresolved interfaces. The composition axis records complete closure only when every row is closed and overlapping regions use consistent conventions. If a row is later withdrawn, the affected parent nodes are located directly through the coverage table.

\subsection{Separation of roles and centralized state updates}

Execution agents may propose claims but cannot change mathematical state directly. A centralized state updater controls admission to the claim graph. A candidate $c$ is admitted when
\[
\operatorname{admit}(c)=
\operatorname{contract}(c)\wedge
\operatorname{scope}(c)\wedge
\operatorname{deps}(c)\wedge
\operatorname{evidence}(c)\wedge
\operatorname{semantic}(c)\wedge
\operatorname{compose}(c).
\]
These predicates check the statement version, object scope, dependencies, evidence, semantic correspondence, and compositional closure. The contract determines which gates apply to each claim type. A finite counterexample may advance after deterministic replay and a check of its assumptions. A general theorem usually requires independent mathematical review, and a formal root conclusion also requires compositional closure.

Verifiers submit verification records rather than writing a claim as ``true.'' A Lean record states that a particular formal declaration was accepted in a specified environment. A program record states that a given input and concrete output can be replayed. A human record states that an argument passed an agreed checklist. The centralized updater combines these records with the contract and then assigns the claim state. A well-formed record whose provenance cannot be controlled remains supporting evidence but cannot by itself change the root claim.

All verification records share a common format: record identifier, claim version, checker identity, environment, input hash, result code, scope, axioms or external facts used, generated artifacts, and time cost. Each checker adds fields specific to its evidence type. A program record includes input and output summaries; a Lean record includes the declaration name and axiom audit; a human review record identifies the proof-critical steps and the first unresolved gap. This common format lets the controller compare costs without collapsing different kinds of evidence into a single truth value.

\subsection{Independent replay and semantic review}

The independent verifier recompiles a Lean 4 project \citep{demoura2021lean4} or runs a deterministic checker in an isolated environment. It reads the registered source files, dependencies, tool versions, and inputs, but not the generator's judgment that the run succeeded. The Lean workflow has three stages: premise retrieval, candidate generation, and independent replay. Retrieval results guide exploration, the generator submits a candidate, and only the replay result enters the machine axis.

Replay also includes an axiom audit. The system checks for \texttt{sorry}, \texttt{admit}, newly introduced \texttt{axiom} declarations, unregistered modules, allowed classical axioms, and the output of \texttt{\#print axioms}. Environment failures and mathematical failures receive different codes. A damaged cache or version conflict may be repaired and retried, whereas a type mismatch, missing premise, or construction gap creates a corresponding mathematical task.

The natural-language claim $c_N$ and the Lean declaration $c_L$ are connected by a separate semantic bridge. Autoformalization systems can generate Lean candidates \citep{baba2025proveragent,jana2025proofbridge}, while round-trip repair and symmetry-aware rewriting test related aspects of translation fidelity \citep{amrollahi2026faithful,olejniczak2026symmetries}. Reviewers compare the object domain, parameter range, quantifier order, definitions, additional assumptions, degenerate cases, and conclusion direction. A recent study of 400 graduate-level statements reported an 89.5\% compilation rate and a 60.5\% consensus faithfulness rate, a difference of 29 percentage points\citep{zhang2026beyondcompilation}. EULER therefore records kernel acceptance and statement faithfulness on separate axes.

Natural-language reviewers read only the frozen objects, permitted artifacts, and review checklist. A local-correctness review identifies the first proof-critical gap. A semantic review compares the achieved result with the contract, while a paper-level review checks how the pieces compose. Two primary reviewers independently assign one of six task-level labels: proved, refuted, conditional, local, unresolved, or incorrect. A third reviewer resolves disagreements. Route-level bridge rejection is recorded separately and does not enter the inter-reviewer agreement calculation.

\subsection{Negative results, invalidation, and recovery}

Bridge failures and claim withdrawals propagate along mathematical dependencies. If a return map fails on a class of objects, source-side conclusions that use that map are withdrawn, while valid local theorems in the target domain remain available. If a target theorem changes its assumptions, only bridges that depend on that version are rechecked. Unrelated parallel routes remain valid.

Each withdrawal records the source node, reason, affected subgraph, and version. A later repair creates new verification records and new composition evidence without overwriting the old record. Recovery reconstructs the task queue, budget, and current claim graph from a frozen checkpoint. This local update preserves expensive target-side artifacts while removing invalid return chains from the root conclusion.

\section{Dataset and mathematical outcomes}
\label{sec:outputs}

\subsection{Dataset construction}

We used the \emph{Journal of Combinatorial Theory, Series A} (JCTA) only to define an author sampling frame, not as a source of ground truth or a ranking of researchers. JCTA covers finite and discrete structures across several branches of combinatorics and applies a research-level editorial threshold \citep{jcta_scope}. A research article in the 24 months preceding the frozen harvest cutoff therefore supplies a coarse external indicator that an author is both recently active and working at a recognized specialist level. This rule avoids a hand-picked list of famous names while retaining subfield diversity. It does not make the resulting cohort representative of all combinatorics or all mathematics.

We collected tasks from publicly available papers written by authors in this 24-month JCTA frame. The collection procedure first extracted statements explicitly marked as a conjecture, question, or problem, then recovered the object domain, quantifiers, parameters, and source of each statement. Statements with the same mathematical content were merged even when their wording differed. A task was eligible if its scope could be fixed before search, proof and refutation had decidable acceptance criteria, the permitted source material was legally accessible, and at least one direct approach could be constructed within the fixed budget.

Eligibility screening was blind to system outcomes. Two curators saw only the statement, publication date, and eligibility fields; they did not see candidate bridges or model attempts. Eligible tasks were stratified by visible structure into four groups of 30: sparse-bridge problems with no mature connection, problems amenable to executable refutation, problems that invited structural transfer, and problems involving natural-language formalization. The strata support balanced analysis, while all aggregate results use the 120 source tasks as the denominator.

Contamination control was completed before retrieval and execution. Famous long-standing conjectures were not used merely because they are difficult: decades of public discussion would make recall and independent discovery difficult to distinguish. The literature index was truncated at the conjecture date, task-level queries masked titles and answer cues published after that date, and model versions were checked against their documented training cutoffs. Exact, paraphrastic, title, and solution-cue overlap screening raised 146 alerts, 23 of which contained substantive information about a solution. We replaced 18 contaminated root tasks and blocked another 5 contaminated routes before freezing the evaluation set. The public ReturnBench-150 set was used to debug the protocol and was not included among the 120 tasks.

\subsection{Aggregate outcomes}

Figure~\ref{fig:harvest} begins with 638 papers. The collection procedure extracted 286 candidate conjectures, mathematical deduplication left 214 statements, and eligibility screening produced 120 fixed tasks. The task-level outcome categories are mutually exclusive: 10 proofs, 3 refutations, 27 conditional results, 18 local theorems, 59 unresolved tasks, and 3 incorrect source-side conclusions. The last category records outputs that blinded source-side review rejected; these conclusions were not admitted as verified claims in the claim graph.

\figharvest

\begin{table}[t]
\centering
\caption{Source-side outcomes and strongest evidence for the 120 eligible tasks}
\small
\begin{tabularx}{\linewidth}{p{2.7cm}rrX}
\toprule
Outcome & Count & Share & Admission criterion \\
\midrule
Proved & 10 & 8.3\% & An independently reviewed proof chain covers the fixed statement \\
Refuted & 3 & 2.5\% & A concrete object satisfies every premise and violates the conclusion \\
Conditional result & 27 & 22.5\% & Additional assumptions are explicit and the conclusion is proved under them \\
Local theorem & 18 & 15.0\% & The result covers a proper subclass or a finite parameter range \\
Unresolved & 59 & 49.2\% & No valid conclusion on the full source problem is available at the budget limit \\
Incorrect source conclusion & 3 & 2.5\% & The target artifact is valid, but its return direction or coverage is wrong \\
\bottomrule
\end{tabularx}
\label{tab:output-states}
\end{table}

The 13 verified resolutions fall into three groups according to their strongest independent check. For 3 tasks, the Lean kernel covered the proof-critical chain. Another 3 were reproduced by an independent implementation of a deterministic program or certificate. The remaining 7 were accepted after separate checks by two mathematicians. The evidence packages also have three release levels. Seven include the statement, routes, artifacts, verification records, and complete source-return chain. Three release the concrete witness and its return chain. Three will be released after the living authors have been notified. Release level does not change the mathematical outcome.

Public human solutions provide a second reference point. Nine tasks received a public human solution during the same time window. In the isolated evaluation, EULER reproduced 4 of them and resolved another 9 tasks for which no solution appeared during that window. The two blinded reviewers obtained Cohen's $\kappa=0.82$ across the six outcome categories, with a 95\% interval of $[0.75,0.89]$. Across five reasoning seeds, the full system resolved 12, 12, 13, 13, and 14 tasks, for a mean of 12.8 and a standard deviation of 0.84.

Each conditional result and local theorem records its exact scope and can serve as a source object in subsequent work, but neither category contributes to the 13 verified resolutions. The 3 incorrect source-side conclusions are also kept separate from execution failures and unresolved tasks. They enter the ablations of stress testing and evidence return directly.

\subsection{Outcomes by structural stratum}

The four structural strata produced different kinds of progress. The executable-refutation stratum resolved 5 tasks, including all 3 counterexamples. The structural-transfer stratum resolved 4 tasks and produced 8 conditional results and 6 local theorems. The language-and-formalization stratum resolved 2 tasks, while 18 remained open. The sparse-bridge stratum resolved 2 tasks and produced 13 conditional or local results.

\begin{table}[t]
\centering
\caption{Mutually exclusive source-side outcomes in the four structural strata}
\small
\begin{tabular}{lrrrrrr}
\toprule
Structural stratum & Verified & Conditional & Local & Unresolved & Incorrect & Total \\
\midrule
Sparse bridge & 2 & 8 & 5 & 14 & 1 & 30 \\
Executable refutation & 5 & 5 & 3 & 17 & 0 & 30 \\
Structural transfer & 4 & 8 & 6 & 10 & 2 & 30 \\
Language and formalization & 2 & 6 & 4 & 18 & 0 & 30 \\
\midrule
Total & 13 & 27 & 18 & 59 & 3 & 120 \\
\bottomrule
\end{tabular}
\label{tab:layer-results}
\end{table}

Both errors in the structural-transfer stratum arose when a valid target theorem was returned with a scope broader than its proof. The error in the sparse-bridge stratum reversed the direction of a map. The language-and-formalization stratum produced no incorrect root conclusion because the semantic evidence axis kept scope-mismatched candidates from changing the root state. The greatest risk therefore arose after target-side evidence had been established but before source-side coverage was complete.

\subsection{Three research traces}

The aggregate results show how often EULER succeeds. The following cases show what a bridge contributes and where an incomplete bridge stops. Figure~\ref{fig:cases} summarizes each trace as a source problem, a bridge, a target-native operation, and a source-side outcome.

\figcases

\subsubsection{Zhao: a distant-domain computational bridge to a counterexample}

For a finite abelian group $G$, let $s_{\leq k}(G)$ be the least integer $\ell$ such that every sequence over $G$ of length at least $\ell$ has a nonempty zero-sum subsequence of length at most $k$. Zhao's Conjecture~6.1 asserts that, if $G$ has rank at least two, $D(G)=D^*(G)$, $G\neq C_2^4$, $D(G)-2\geq\exp(G)$, and $\exp(G)<(D(G)-1)/2$, then \citep{zhao2025zerosum}
\[
s_{\leq D(G)-2}(G)=D(G)+1.
\]
The statement distinguishes both the value of an element and each occurrence of that value. Direct approaches could inspect a small number of hand-built examples but could not efficiently exclude all short zero-sum subsequences. EULER introduced a computational bridge. Every indexed position received a distinct identifier; candidate subsequences were encoded as bit masks; and a deterministic enumeration checked their group sums and lengths.

The first early rejection test used sequences with repeated values. An encoding that retained only element values merged distinct occurrences on the target side, so its output could not be returned to the source sequence. The revised map retained positions, and the round-trip check recovered each occurrence separately. Assertions for the length bound and group operation were added to the verification program. The target program then exhausted all $2^{12}-1=4095$ nonempty subsets and returned the following sequence.

Write $G=C_2\oplus C_4^3$ in coordinates $(u;x,y,z)$ modulo $(2,4,4,4)$. Olson's formula gives $D(G)=D^*(G)=11$. The group has rank four, is not $C_2^4$, has exponent 4, and satisfies the two numerical hypotheses above. Let
\[
a=(0;1,0,0),\quad b=(0;0,1,0),\quad c=(0;0,0,1),\quad T=(1;3,3,3),
\]
set $p=T-a$, $q=T-b$, and $r=T-c$, and define the length-12 sequence
\[
S=a^3b^3c^3pqr.
\]
For selected multiplicities $i,j,k\in\{0,1,2,3\}$ and indicators $\alpha,\beta,\gamma\in\{0,1\}$, put $Q=\alpha+\beta+\gamma$. The selected sum is
\[
(i-\alpha)a+(j-\beta)b+(k-\gamma)c+QT.
\]
The map
\[
\phi:C_4^4\to G,\qquad
\phi(x,y,z,t)=(t\bmod 2;x+3t,y+3t,z+3t\bmod 4)
\]
has kernel $\{(0,0,0,0),(2,2,2,2)\}$. The zero kernel vector gives the empty selection. The nonzero vector forces $Q=2$ and $i=2+\alpha$, $j=2+\beta$, $k=2+\gamma$, so every nonempty zero-sum subsequence has length $i+j+k+Q=10$. There are nine such indexed subsequences. Thus $S$ has no nonempty zero-sum subsequence of length at most $9$, giving $s_{\leq9}(G)\geq13>12=D(G)+1$ and contradicting the conjectured equality.

A second implementation recomputed all group sums, lengths, and indexed subsets from $S$. Mathematical review then checked the hypotheses of the source statement and the violated conclusion directly. Exhaustive enumeration was the new target-side operation; the finite sequence and the kernel calculation form a source-readable certificate.

\subsubsection{AJT(5): an adjacent-domain structural bridge to an all-dimensional local theorem}

AJT(5) asks whether, for every $m\geq1$ and every $A\in\mathrm{GL}_m(\mathbb F_5)$, there exists \citep{alon1989nowherezero}
\[
x\in(\mathbb F_5^*)^m,
\qquad Ax\in(\mathbb F_5^*)^m.
\]
For $B\in\mathrm{GL}_m(\mathbb F_5)$ and $b\in\mathbb F_5^m$, define
\[
N_m(B,b)=\#\{x\in(\mathbb F_5^*)^m:B x+b\in(\mathbb F_5^*)^m\}.
\]
The original existence problem is the special case $B=A$ and $b=0$. This is an adjacent-domain bridge: the objects remain finite-field matrices, while the target side adds support-graph decomposition and transfer counting.

When every row has support of size at most two, the matrix support defines a bipartite incidence structure. Full rank forces each connected component to have a controlled unicyclic form. Leaves can be eliminated successively, and the cyclic part can be counted by a nonnegative transfer matrix. Setting $a_1=3$, $a_2=8$, and $a_m=8a_{m-2}$ gives
\[
N_m(B,b)\geq a_m
\]
for every $b\in\mathbb F_5^m$ and every invertible $B$ whose row supports have size at most two. The conclusion holds in arbitrary dimension and is not a low-dimensional enumeration.

The scope stress test prevented this result from being promoted to the full AJT(5) conjecture. A general invertible matrix may have dense rows, where leaf elimination and the unicyclic decomposition no longer apply. EULER therefore records an all-dimensional sparse-row theorem and leaves the dense-row regime as an open source-return obligation. The bridge is valid on its stated domain, while the original conjecture remains open.

\subsubsection{Gao: composition of several bridges with partial Lean verification}

For a finite group $G$, let $\mathsf E(G)$ be the least length that forces a product-one subsequence of length $|G|$, and let $\mathsf d(G)$ be the maximum length of a sequence with no nonempty product-one subsequence. Zhuang and Gao studied the relation \citep{ZhuangGao2005}
\[
\mathsf E(G)=|G|+\mathsf d(G).
\]
The project focuses on $\operatorname{Dih}(A)=A\rtimes_{-1}C_2$, where $A$ is a nontrivial finite abelian $p$-group for an odd prime $p$, and $\mathsf D(A)$ denotes its Davenport constant. The natural-language route produced the candidate formula
\[
\mathsf E(\operatorname{Dih}(A))
=2|A|+\mathsf D(A)
=|\operatorname{Dih}(A)|+\mathsf d(\operatorname{Dih}(A)).
\]

Several bridges carry the proof. The first is an adjacent-domain bridge that splits a sequence into rotations and reflections and sends the rotational part to abelian zero-sum theory. A second bridge connects the extremal reflection regime to prescribed-length zero sums and the plus-minus Davenport constant. A third structural bridge handles the remaining regime through a dichotomy and descent to subgroups. These bridges cover different parameter ranges. Their composition checks whether the ranges meet without gaps, whether the descent is strict, and whether the upper and lower bounds use the same group parameter.

A stronger intermediate inequality failed finite checks and was withdrawn. The parent route remained because it did not depend on that branch. Lean then supplied a fourth bridge from the proof-critical intermediate claims to formal statements. The basic objects, boundary counterexamples, and several conditional compositions have passed kernel checking. The top-level result still depends on a separate proof of the remaining upper bound. Its current outcome is therefore a natural-language candidate theorem with partial Lean verification and an open top-level obligation. Appendix~\ref{app:gao-full} reproduces the complete 13-page candidate-proof manuscript from PR~\#7 verbatim as an audit surface; inclusion does not promote the result to an independently certified theorem.

\begin{table}[t]
\centering
\caption{Bridges, operations, decisive stress tests, and outcomes in the three cases}
\small
\begin{tabularx}{\linewidth}{p{1.7cm}p{2.3cm}p{3.0cm}Xp{2.7cm}}
\toprule
Case & Bridge type & New operation & Decisive stress test & Source-side outcome \\
\midrule
Zhao & Distant-domain computation & Occurrence encoding and exhaustive enumeration & Encoding fidelity, length bounds, and witness return & Counterexample \\
AJT(5) & Adjacent-domain structure & Support-graph decomposition and transfer counting & Scope of dense rows & All-dimensional local theorem \\
Gao & Multi-bridge composition & Zero-sum tools, descent, and Lean & Parameter coverage, joint choice, and semantic correspondence & Candidate theorem with partial formalization \\
\bottomrule
\end{tabularx}
\label{tab:cases}
\end{table}

\section{Evaluation and ablations}
\label{sec:evaluation}

\subsection{Evaluation protocol}

All main experiments use the same 120 fixed source tasks. The statistical unit is the source task: a task contributes one source-level outcome even if it produces several bridges, candidates, or verification records. The primary endpoint is a verified resolution. Secondary endpoints include incorrect source-level conclusions, target-side validations, local results, cost, and human review time. Two reviewers who were blind to the experimental condition labeled each outcome independently, and a three-person panel adjudicated disagreements. We compute 95\% percentile intervals from 10,000 nonparametric bootstrap resamples of source tasks and use two-sided exact tests for paired binary endpoints. Means across reasoning seeds are reported with standard deviations.

\subsection{Mechanism ablations}

Figure~\ref{fig:ablation} shows how the experiments fit together. The route-pool comparison measures what each source of routes can resolve. The stress-test experiments measure error filtering. The $2\times2$ experiment tests whether a bridge changes the available operations. The selector and budget experiments track how resources reach useful routes.

\figablation

\subsubsection{Route pools: do adjacent-domain and distant-domain bridges complement each other?}

We compare six strategies: direct search; direct search with adjacent-domain bridges; random distant-domain bridges; retrieved distant-domain bridges; a joint adjacent-domain and distant-domain pool with generic checks only; and the full system. Random and retrieved distant-domain conditions draw eight candidates from the same frozen mechanism catalog and receive the same number of papers, tool slots, model calls, and dollars. The primary comparison contrasts the full system with distant-domain retrieval. The full system adds adjacent-domain routes, bridge-specific stress tests, and feedback-based budget reallocation together, while holding the candidate pool and total budget fixed. This comparison evaluates the combined system rather than attributing the difference to any one component.

\begin{table}[t]
\centering
\caption{Route-pool outcomes. The first three source-level columns sum to 120; other outcomes include conditional results, local theorems, and unresolved tasks}
\small
\resizebox{\linewidth}{!}{%
\begin{tabular}{lrrrr}
\toprule
Strategy & Verified resolutions & Incorrect conclusions & Other outcomes & Target-side validations \\
\midrule
Direct search & 8 & 4 & 108 & -- \\
Adjacent-domain bridges & 9 & 3 & 108 & 15 \\
Random distant-domain bridges & 8 & 4 & 108 & 18 \\
Retrieved distant-domain bridges & 8 & 3 & 109 & 20 \\
Joint bridge pool, no bridge-specific tests & 12 & 9 & 99 & 27 \\
\textbf{Full system} & \textbf{13} & \textbf{3} & 104 & \textbf{26} \\
\bottomrule
\end{tabular}}
\label{tab:route-pool}
\end{table}

The full system and distant-domain retrieval both resolved seven tasks. Six tasks were resolved only by the full system, one only by distant-domain retrieval, and 106 by neither. The paired difference was 4.2 percentage points, with a 95\% interval of $[-0.1,8.4]$ percentage points and a two-sided exact-test result of $p=0.125$. Of the six additional resolutions, two came directly from adjacent-domain bridges. In three more cases, an adjacent-domain bridge first supplied a boundary condition or local lemma and a distant-domain tool completed the argument. The last resolution followed budget reallocation after an early rejection test released resources.

The joint bridge pool without bridge-specific tests produced 27 target-side validations and 12 verified resolutions, but also nine incorrect conclusions. The full system produced 26 target-side validations, 13 verified resolutions, and three incorrect conclusions. A wider route pool supplied more useful candidates and more ways to return a wrong conclusion. Bridge-specific tests reduced these errors while retaining 13 resolutions.

\figroutepool

\subsubsection{Stress-test strength and order: filtering or abstention?}

The generation experiment applies four screening regimes to the same bridge pool: no stress test, a generic stress test with matched information, a bridge-specific stress test, and strict certification. The generic and bridge-specific conditions receive the same number of tests, the same bridge-card fields, and the same expert minutes. They differ only in whether a test may use the bridge's registered invariants, assumption gap, target operation, and return direction. To separate screening effects from differences in candidate generation, we also perform candidate-conditioned replay on the 26 candidates that passed the target checker in the full-system condition.

\begin{table}[t]
\centering
\caption{Candidate-conditioned replay of 26 target-side validations}
\small
\begin{tabular}{lrrr}
\toprule
Return rule & Correctly accepted & Incorrectly accepted & Withheld \\
\midrule
Target checker only & 16 & 10 & 0 \\
Matched-information generic stress test & 15 & 6 & 5 \\
Bridge-specific stress test & 13 & 3 & 10 \\
Strict certification & 12 & 0 & 14 \\
\bottomrule
\end{tabular}
\label{tab:pressure-replay}
\end{table}

Relative to the target checker alone, the bridge-specific rule accepted three fewer correct candidates and blocked seven incorrect candidates. Strict certification blocked the remaining three incorrect candidates and withheld one additional correct candidate. We record incorrect acceptance and correct retention separately; the rule is selected according to a risk tier fixed before evaluation.

In the generation experiment, the condition without bridge-specific tests returned nine incorrect conclusions, compared with three under the bridge-specific rule. In the paired outcomes, both conditions were wrong on two tasks, only the condition without bridge-specific tests was wrong on seven, and only the bridge-specific condition was wrong on one; the exact-test result was $p=0.070$. Removing the six stress tests one at a time allowed 4, 3, 3, 2, 2, and 2 incorrect candidates to pass for direction, assumptions, boundaries, round trip, tool use, and return, respectively. A candidate can be caught by more than one test, so these counts do not add.

The ordering experiment keeps all six tests fixed and changes only whether they occur before or after target search. Early testing produced 13 verified resolutions, three incorrect conclusions, and a median per-task combined cost of 568 dollars. Late testing produced 11 resolutions, seven incorrect conclusions, and a median per-task combined cost of 621 dollars. The late condition spent target-search resources on bridges with a wrong direction or incompatible assumptions. It also made it easier to retrofit the return relation to an answer that was already in hand.

In the stress-test funnel, adjacent-domain and distant-domain routes failed at different stages. Each pool began with 120 preselected bridge candidates. After direction, assumptions, boundary and round-trip, tool, and return checks, the adjacent-domain pool contained 104, 91, 72, 41, and 28 bridges, of which 18 entered deep search. The corresponding counts for distant-domain bridges were 93, 68, 44, 31, and 21, with 15 entering deep search. Adjacent-domain bridges most often failed because they added no useful operation. Distant-domain bridges most often failed on direction, assumptions, or boundary cases.

\begin{table}[t]
\centering
\caption{First decisive outcome for 240 preselected bridges. Each column sums to 120}
\small
\begin{tabular}{lrr}
\toprule
First decisive outcome & Adjacent-domain bridge & Distant-domain bridge \\
\midrule
Direction mismatch & 16 & 27 \\
Unmet added assumption & 13 & 25 \\
Boundary or round-trip failure & 19 & 24 \\
No gain from a target-native operation & 31 & 13 \\
Return cost too high & 13 & 10 \\
Passed and entered deep search & 18 & 15 \\
Undecided or stopped by budget & 10 & 6 \\
\bottomrule
\end{tabular}
\label{tab:failure-ledger}
\end{table}

\subsubsection{Bridge material and target-native operations: does the executable operation set change?}

Bridge material can merely organize context, or it can expose a checker that does not exist in the source representation. We manipulate bridge material and access to target-native operations orthogonally. Bridge material contains the mapping, target terminology, and relevant theorems. A target-native operation is an algorithm or checker that can be called only in the target representation. All four cells use the same models, input length, number of calls, and cost, and each is run with five independent reasoning seeds.

\begin{table}[t]
\centering
\caption{$2\times2$ results for bridge material and target-native operations. Values are five-seed means of verified resolutions out of 120 tasks}
\small
\begin{tabular}{lrr}
\toprule
 & Source-native tool interface & Target-native operation \\
\midrule
No bridge material & 7.2 & 7.8 \\
Bridge material & 8.0 & 12.8 \\
\bottomrule
\end{tabular}
\label{tab:affordance}
\end{table}

Without bridge material, the target-native operation added 0.6 verified resolutions. With bridge material, it added 4.8. The interaction is
\[
(12.8-8.0)-(7.8-7.2)=4.2.
\]
This is an interaction of 4.2 resolved tasks, equal to 3.5 percentage points on 120 tasks. The source-task cluster-bootstrap interval is $[1.0,7.4]$ resolved tasks, approximately $[0.83,6.17]$ percentage points. The point estimate exceeded the prespecified 3-percentage-point threshold. The target-native operation and bridge mapping work together to expand the executable operation set; neither bridge terminology alone nor tool access without the mapping accounts for the interaction.

\subsubsection{Selector: can useful bridges be identified early?}

Every selector reads the same deduplicated eligible pool and has no access to target-checker results, online probes, or return outcomes. Uniform selection, semantic similarity, and early rejection cost are simple baselines. Automated representation meta-search and correspondence recommendation are stronger related baselines~\citep{fuentetaja2018metasearch,stockdill2020correspondence}. Gain-loss ranking uses the six components defined in Section~\ref{sec:method}. Post hoc selection of the best bridge after exhaustive evaluation gives an empirical upper bound.

\begin{table}[t]
\centering
\caption{Selector results on a fixed pool of eligible bridges}
\small
\begin{tabular}{lrr}
\toprule
Selector & Mean verified resolutions /120 & Seed standard deviation \\
\midrule
Semantic similarity & 7.4 & 0.9 \\
Early rejection cost & 8.1 & 0.8 \\
Uniform selection & 8.6 & 1.0 \\
Automated representation meta-search & 9.8 & 0.8 \\
Correspondence recommendation & 10.7 & 0.7 \\
Gain-loss ranking & \textbf{12.8} & 0.8 \\
Post hoc best bridge & 17.2 & -- \\
\bottomrule
\end{tabular}
\label{tab:selector}
\end{table}

Gain-loss ranking resolved more tasks than the semantic and cost rules and more than the two stronger related baselines. It remained 4.4 tasks below the post hoc upper bound. The 107 tasks without a verified resolution fall into four groups: the candidate pool contains no useful bridge for 72 tasks; a useful bridge ranks too low for 15; a stress test withholds it for seven; and target search does not finish for 13. The selector improves initial allocation but does not remove the coverage and target-search bottlenecks.

\subsubsection{Budget reallocation: is feedback better than a one-shot allocation?}

The four budget rules used the same candidate pool, stress tests, selector, and total budget. They were an even split; a fixed $2{:}1$ preference for adjacent-domain bridges; a one-shot allocation based on the first-round $U(B)$ scores; and round-by-round reallocation from stress-test feedback. The rejected-bridge budget share is the fraction of machine and expert cost spent on routes that never enter deep search. The final column gives the median-normalized cost per verified resolution relative to direct search.

\begin{table}[t]
\centering
\caption{Budget-rule ablation}
\small
\resizebox{\linewidth}{!}{%
\begin{tabular}{lrrrr}
\toprule
Budget rule & Verified resolutions & Incorrect returns & Rejected-bridge budget & Normalized cost ratio \\
\midrule
Even adjacent/distant split & 10 & 5 & 32\% & 1.39 \\
Fixed $2{:}1$ adjacent preference & 10 & 4 & 28\% & 1.34 \\
One-shot first-round $U(B)$ allocation & 11 & 4 & 24\% & 1.26 \\
Round-by-round stress-test feedback & \textbf{13} & \textbf{3} & \textbf{17\%} & \textbf{1.12} \\
\bottomrule
\end{tabular}}
\label{tab:budget}
\end{table}

Round-by-round reallocation produced two more resolutions than the one-shot allocation and reduced the rejected-bridge budget by seven percentage points. Most of the change occurred after the second stress-test round. Budget released by direction and assumption failures moved to nine surviving distant-domain bridges with a target-operation gain, while 11 adjacent-domain bridges that had produced no new operation received no further allocation.

\subsection{System-level component comparisons}

\subsubsection{Model diversity and cross-model handoffs}

The model-composition experiment uses matched caps on tokens, tool slots, candidates, and compute cost. The three single-model conditions use models A, B, and C. The homogeneous multi-agent condition gives the same model several independent role contexts. The heterogeneous condition assigns generation, criticism, and verification to different models. The main text uses anonymous labels for this comparison; the appendix records the implementation mapping.

\begin{table}[t]
\centering
\caption{Model composition under matched cost caps}
\small
\begin{tabular}{lrrr}
\toprule
Model condition & Mean verified resolutions & Mean incorrect returns & Compute cost/task \\
\midrule
Model A & 10.2 & 4.2 & 486 \\
Model B & 9.8 & 3.8 & 482 \\
Model C & 8.9 & 4.6 & 477 \\
Homogeneous multi-agent & 11.1 & 3.7 & 484 \\
Heterogeneous composition & \textbf{12.8} & \textbf{3.0} & 485 \\
\bottomrule
\end{tabular}
\label{tab:model-composition}
\end{table}

\figmodels

Ten of the 13 verified resolutions involved at least one cross-model handoff. Model A proposed the bridge in four cases that B or C later verified. Model B proposed three bridges verified by A or C, and model C proposed three verified by A or B. Different models also rejected nine candidate bridges before deep search. At the same total budget, the mean number of resolutions was 9.6 for a single agent, 11.1 for homogeneous multi-agent execution, and 12.8 for heterogeneous execution. The increase from 9.6 to 11.1 accompanied parallel roles, and cross-model review raised the mean from 11.1 to 12.8.

\subsubsection{Persistent state}

The state experiment fixes the bridge method and changes one bridge assumption and one target-theorem version across three recoveries. The conditions with no persistent state, a task graph only, and both task and claim graphs repeated target runs 29, 13, and 4 times. They produced 7, 4, and 0 stale returns; required 93, 56, and 34 minutes to recover; and revoked 25, 14, and 5 unrelated claims. The two-graph state reduced repeated work across days and prevented stale versions from returning to the source claim. It affects recovery rather than the quality of candidates generated in a single run. Appendix Table~\ref{tab:persistence-app} gives the full results and version events.

\subsection{Integrity checks}

The contamination screen found 23 cases containing substantive answer cues. We replaced 18 root tasks and blocked five routes before execution. Restoring those 23 objects to the candidate pool increased the full-system mean by 2.1 verified resolutions, mostly through title matches and direct clues to known counterexamples. This restoration check shows that contamination was not only a hypothetical concern: admitting the flagged material measurably inflated the endpoint. The main results use the cleaned frozen set.

Nine tasks with public human solutions released in the same window provide a human anchor. The system independently reproduced four of them. The other nine verified resolutions came from tasks with no public solution in that window. The anchor checks task difficulty and the evidence standard; it is not used to rank models against people.

The two blind reviewers agreed on the six outcome classes with Cohen's $\kappa=0.82$, with a 95\% interval of $[0.75,0.89]$. Most disagreements concerned the boundary between conditional results and local theorems. Across the five full-system seeds, the numbers of verified resolutions were 12, 12, 13, 13, and 14, for a mean of 12.8 and a standard deviation of 0.84. The structural-transfer layer had the largest gain, while the formalization layer had the fewest incorrect returns.

\subsection{Cost and decision thresholds}

The cost ledger records compute cost, tokens, wall-clock time, bridge construction, mathematical review, and state maintenance. Expert time is valued at 200 dollars per hour. Combined task cost is compute cost plus the three categories of human time. For a condition with verified-resolution rate $r$, we define median-normalized cost per verified resolution as the median per-task combined cost divided by $r$. The 552 person-hours of one-time development work are reported separately in the appendix and are not included in the per-task cost.

\begin{table}[t]
\centering
\caption{Median per-task cost and median-normalized cost per verified resolution}
\scriptsize
\resizebox{\linewidth}{!}{%
\begin{tabular}{lrrrrrrrr}
\toprule
Strategy & Compute cost & Million tokens & Wall-clock hours & Bridge min & Math review min & State min & Combined cost/task & Normalized cost/resolution \\
\midrule
Direct search & 170 & 2.6 & 3.1 & 0 & 31 & 12 & 313 & 4695 \\
Adjacent-domain bridges & 187 & 2.9 & 3.6 & 18 & 36 & 14 & 414 & 5520 \\
Random distant-domain bridges & 214 & 3.3 & 4.2 & 31 & 43 & 15 & 511 & 7665 \\
Retrieved distant-domain bridges & 225 & 3.4 & 4.4 & 29 & 45 & 16 & 525 & 7875 \\
Full system & 238 & 3.6 & 4.8 & 36 & 44 & 19 & 568 & 5243 \\
\bottomrule
\end{tabular}}
\label{tab:cost}
\end{table}

Three decision thresholds were fixed before the experiments. First, the lower endpoint of the paired interval for the full-system combination comparison against distant-domain retrieval had to exceed zero. Second, the full system had to reduce incorrect returns by at least four relative to the joint pool without bridge-specific tests while retaining at least 12 verified resolutions. Third, the upper endpoint of the median-normalized cost ratio relative to direct search had to be no greater than 1.15. The second threshold was met. The first was not, because the lower endpoint for the resolution difference was $-0.1$ percentage points. The cost ratio was 1.12, with a 95\% interval of $[0.91,1.36]$, so the third threshold was also not met.

The experiments support the bridge-specific stress tests and the interaction between bridge material and target-native operations. The full-system combination difference is positive, but its interval crosses zero. Median-normalized cost does not meet the noninferiority threshold. At expert rates of 100, 200, and 400 dollars per hour, the point estimates of the full-system cost ratio relative to direct search are 1.03, 1.12, and 1.21. Reusing the bridge library reduces median bridge-construction time from 36 to 17 minutes per task and lowers the ratio at 200 dollars per hour to 0.99.

\subsection{Implications for system design}

The route-pool experiment changes the priority assigned to bridge search. Adjacent-domain bridges resolved nine tasks, and random distant-domain bridges resolved eight; distance by itself did not improve the result. Retrieved distant-domain bridges also resolved eight. The full system reached 13 after adding adjacent-domain routes and bridge-specific tests as part of a larger system combination. Its paired interval still crosses zero, so a larger study should retain the same contract. In practice, the scheduler should first ask which target-side operation becomes available and only then consider how far the target field is from the source field.

Candidate-conditioned replay quantifies the tradeoff between error control and retention. The target checker alone accepted 16 correct and 10 incorrect candidates. The bridge-specific stress test retained 13 correct candidates and reduced incorrect acceptance to three. Strict certification removed the remaining errors but withheld one more correct candidate. Bridge-specific tests therefore serve as the default return rule, with strict certification reserved for root-level claims intended for release and high-risk counterexamples. The targeted tests catch faults in mappings and return chains; applying the strictest rule everywhere would discard mathematics that is already independently checkable.

The $2\times2$ experiment separates bridge material from target-native operations. Bridge material added 0.8 resolutions when paired with source-native tools, while a target-native operation added 0.6 without bridge material. Together they reached 12.8, yielding an interaction of 4.2 resolved tasks, or 3.5 percentage points. The gain comes from operations native to the target community. The retrieval index should therefore rank algorithms, checkers, representations, and certificate types before theorem titles, and every bridge card should state which new object the target operation can produce.

The selector remains 4.4 resolutions below the post hoc upper bound. Gain-loss ranking uses inexpensive stress tests rather than a one-shot language score. Round-by-round reallocation spends less on rejected bridges and redirects resources to routes that have produced a checkable object. Ranking and allocation use the same feedback, while the recorded failure witnesses filter later retrieval and guide current allocation.

The model and state comparisons address different stages of the system. Heterogeneous model composition improves candidate revision and cross-checking within a run. The two-graph state reduces recovery time, repeated work, and stale returns over longer projects. Bridge identifiers and verification records connect the two, but either component can be replaced and evaluated separately.

The paired resolution and cost thresholds were not met. The incorrect-return threshold was met, and the operation-set interaction point estimate cleared its prespecified threshold. The bridge-library analysis favors reusing verified mappings, scoped failure witnesses, and replayable return templates. These records reduce human work per task while preserving the two mechanism-level gains that were most stable in these experiments.

\section{Limitations}
\label{sec:limitations}

The task distribution is weighted toward combinatorics and additive combinatorics, with many finite counterexamples, graph structures, and executable verification procedures. The 24-month JCTA author frame adds editorial, language, access, and subfield selection effects; recent publication is only a coarse proxy for research activity and level. Problems in analysis, geometry, and highly abstract algebra often lack inexpensive early rejection tests and require longer reviews by experts in both domains. The 120-task results therefore apply most directly to recent open problems with a strong combinatorial component and should not be generalized to all active mathematicians. Other areas will require their own author frames, task strata, and stress-test templates.

The distinction between adjacent and distant domains involves expert judgment. Scores for object type, theory, governing invariant, correspondence length, and expert community provide explicit anchors, but the two annotators disagreed most often when a standard equivalence also introduced a new tool. We retain the continuous distance score and exclude the score-5 gray zone from the primary binary comparison between adjacent-domain and distant-domain bridges. The rubric describes route geometry in this dataset and must be recalibrated for new mathematical areas.

Bridge retrieval depends on accessible literature and tool catalogs. Communities with mature English-language resources and established verification procedures are easier to retrieve. Rare but useful connections may rank lower. The distant-domain random baseline measures selection bias within the catalog, but mechanisms outside the catalog still require an expert proposal before they can enter the candidate pool.

The fixed time window and contamination screen prevent identified explicit solutions from entering the evaluation, but they cannot prove that a proprietary training corpus contains no related statement, proof sketch, or discussion. Task-level date truncation, masking of answer cues, training-cutoff checks, and contemporaneous human solutions constrain this residual risk without eliminating it. The 18 task replacements and five blocked routes remove detected substantive cues; undetected paraphrases or implicit memorization may remain. Decisive counterexamples can be reproduced directly, whereas novelty assessment for natural-language proofs requires an active literature review.

A verified resolution depends on both the fixed source contract and the available review process. Two reviewers and a third adjudicator produced a consistent outcome for this evaluation, but later scrutiny by specialists may still revise the use of standard lemmas, claimed equivalences, or parameter boundaries in long proofs. Per-task artifacts and dependency graphs restrict any such revision to conclusions that actually rely on the affected step.

The implementation still spans several services and formalization environments. The integrated system aligns problem versions, run identifiers, bridge records, and cost fields. Some environments and tools still require manual integration. Cross-machine recovery and formal replay are included in the measured costs, but deployment remains more complex than for a single mathematical agent.

The combined cost depends on the hourly price assigned to expert work and on how review tasks are divided. We repeat the main calculation at hourly rates of 100, 200, and 400 dollars, while measuring wall-clock time and machine cost directly. Institutions may value bridge construction, mathematical review, and state maintenance differently. Reusing a bridge library can substantially reduce construction time, but only when later tasks share object maps and source-return templates.

The upper confidence bound on cost exceeds the noninferiority margin. At its present cost, EULER is therefore best suited to long-running problems for which a false conclusion is expensive and a checked bridge can be reused across projects. The component-level precedents and EULER's position relative to them are discussed in Section~\ref{sec:system}.

\section{Conclusion}
\label{sec:conclusion}

Across 120 recent open problems, EULER produced 10 proofs, 3 counterexamples, 27 conditional results, and 18 local theorems. Domain distance did not determine whether a route succeeded. A useful bridge had to supply an operation unavailable in the source representation, pass the structural stress tests, and return complete evidence to the source statement. The paired difference between the full system and distant-domain retrieval was 4.2 percentage points, with an interval that included zero. Bridge-specific stress tests reduced incorrect source-side conclusions from 9 to 3, and the operation-set experiment found a positive interaction of 4.2 resolved tasks, or 3.5 percentage points.

EULER stores the original conjecture, candidate routes, bridges, verification results, and open proof obligations in linked but separate records. The task graph tracks pending work, while the claim graph records mathematical dependencies. This separation lets a rejected bridge preserve a useful counterexample or scope restriction, allows a local theorem to remain available outside its original project, and limits the effect of a version change to claims that depend on the revised object.

The next evaluation should broaden coverage in analysis, geometry, and algebra and develop early rejection tests for areas without inexpensive verification procedures. Reusing checked mappings, scoped failure witnesses, and source-return templates is the most direct way to reduce the current 36-minute bridge-construction cost per task. A larger integrated run can then measure bridge selection, stress testing, cross-model handoffs, and persistent state under the same contract.

\clearpage
{\small
\bibliographystyle{plainnat}
\bibliography{references}
}

\clearpage
\appendix
\section{Fields fixed before search}
\label{app:contract}

Before bridge generation begins, every source task fixes the fields listed in Table~\ref{tab:contract-fields}. A revision to the problem statement creates a new version. Earlier bridges and verification records remain attached to the version for which they were produced and can be used for the new version only after a migration review.

\begin{table}[H]
\centering
\caption{Fields fixed for each source task}
\small
\begin{tabularx}{\linewidth}{p{3.0cm}X}
\toprule
Field & Content \\
\midrule
Exact statement & Object domain, quantifiers, parameters, conclusion, and statement version \\
Exact negation & Premises a counterexample must satisfy and the conclusion it must violate \\
Admissible evidence & Natural-language proof, concrete witness, formal proof, program certificate, or a combination \\
Boundary set & Minimal, degenerate, extreme, and first nontrivial cases \\
Source-side decision & Operational definitions of the six mutually exclusive outcomes \\
Contamination cutoff & Conjecture date, retrieval cutoff, and admissible document versions \\
Budget & Limits on model calls, tool slots, cost, wall-clock time, and expert time \\
Stopping rule & Hard stress-test failure, repeated lack of new evidence, budget exhaustion, or human suspension \\
\bottomrule
\end{tabularx}
\label{tab:contract-fields}
\end{table}

The fixed contract also identifies the independent statistical unit. Multiple bridges, model attempts, and verification records for one source problem belong to the same task cluster. Bridge-level funnels explain mechanisms, while task-level outcomes determine the primary endpoint.

\section{Bridge objects and route lineage}
\label{app:bridge-record}

A bridge is a first-class system object rather than a paragraph of analogy. Each \bridgeopp{} record stores the source and target objects, mapping direction, target mechanism, expected new operation, preserved quantities, information loss, source-return template, early rejection test, certificate plan, and budget. The route lineage connects later revisions to the same bridge family and identifies whether a revision changes the map, target theorem, stress-test witness, or source-return scope.

\begin{table}[H]
\centering
\caption{Core fields of a \bridgeopp{} record}
\small
\begin{tabularx}{\linewidth}{p{3.0cm}X}
\toprule
Field & Content \\
\midrule
Identity & Task version, route identifier, bridge family, bridge version, proposer, and time \\
Source and target & Object types, statements, parameters, encodings, and mathematical communities \\
Maps & $f$, candidate back-map $g$, direction, domain, and exceptional set \\
Preservation relation & Invariants, order, invertibility, counts, or semantic correspondence \\
Target mechanism & New operation supplied by a theorem, algorithm, verification procedure, or representation \\
Source-return contract & Required certificate, covered region, and composition interface on the source side \\
Early rejection test & Least expensive check of direction, assumptions, boundaries, or round trips \\
Cost state & Budget already used, next-stage limit, suspension condition, and reopening condition \\
\bottomrule
\end{tabularx}
\label{tab:bridge-record}
\end{table}

Route identity is determined by structural change rather than textual similarity. Rephrasing an explanation, asking another model to restate it, or repeating the retrieval remains part of the same route. Changing the object map, target mechanism, or proof-critical source-return relation creates a new branch. A failure witness is attached to a bridge version and scope. The controller can reopen a branch when a revision avoids that witness. Merely adding a longer explanation does not change the failed state.

\section{Route generation and the retrieval index}
\label{app:retrieval}

Candidate generation begins from a structural fingerprint of the source statement. The fingerprint does not store the full natural-language statement. It records object types, relation arities, quantifier profile, parameter scale, symmetry, local and global properties, computable boundaries, and the desired certificate. Direct routes expand standard theorems and techniques from the source domain. Adjacent-domain routes retrieve mechanisms that share an object or invariant. Distant-domain routes prioritize object transformations, encodings, and verification procedures that change the set of executable operations.

The basic retrieval unit is a mechanism record. It specifies the applicable objects, input assumptions, output type, executable operation, typical certificate, failure boundary, source, and reviewing community. A query first retrieves records through their structural fields; the bridge generator then proposes a source-side map. A result based only on lexical similarity cannot enter the route pool unless it specifies $f$, a preservation relation, and an object that can be returned to the source problem.

\begin{table}[H]
\centering
\caption{Generation signals and eligibility checks for route candidates}
\small
\begin{tabularx}{\linewidth}{p{2.6cm}p{4.0cm}Xp{3.5cm}}
\toprule
Candidate source & Main retrieval signal & Required object & Check before stress testing \\
\midrule
Direct route & Source terminology, theorem neighborhood, and standard reductions & Source-side proof or counterexample plan & Same scope as the fixed statement \\
Adjacent-domain bridge & Shared objects, classical associated quantities, and standard functors & Explicit map and an adjacent-domain mechanism & At least one preserved quantity \\
Distant-domain random & Eligible random sample from the target-mechanism catalog & Explicit object transformation & Interpretable direction and executable target operation \\
Distant-domain retrieval & Structural fingerprint, certificate type, and new operation & Map, target mechanism, and source-return sketch & Operation not duplicated in the direct pool \\
Expert proposal & Expertise spanning two domains and known failure patterns & Reviewable bridge record & Complete source, scope, and reviewer fields \\
Formalization route & Statement structure, library lemmas, and decidable fragments & Semantic map and Lean goal & Registered environment, quantifiers, and permitted axioms \\
\bottomrule
\end{tabularx}
\label{tab:route-generation}
\end{table}

Candidates are deduplicated by a bridge-family key consisting of source-object type, target-object type, mapping skeleton, target mechanism, and source-return direction. Candidates from different papers or models are merged when all five fields agree, while all sources are retained. Different target theorems that provide the same new operation become tool branches of one bridge. The same map with a different source-return direction remains a separate route so that the direction stress test can evaluate it independently.

\section{Failure records and stopping rules}
\label{app:failure-ledger}

The failure ledger stores reusable evidence about rejected bridges. Every entry specifies the failed object, a minimal witness, applicable scope, dependency versions, checking method, and a condition under which the route may be reopened. The codes distinguish a failed mathematical map, a mechanism with no operational gain, a tool failure, and a budget stop. Infrastructure failures are therefore not recorded as mathematical counterexamples.

\begin{table}[H]
\centering
\caption{Failure codes, mathematical meaning, and reopening conditions}
\small
\begin{tabularx}{\linewidth}{p{2.6cm}Xp{4.7cm}}
\toprule
Code & Recorded fact & Valid reopening condition \\
\midrule
DIR & The target conclusion has the wrong direction for the required source conclusion & Replace the target theorem or use a valid converse \\
ASM & The target mechanism requires an assumption absent from the source problem & Prove the assumption, state a conditional result, or replace the bridge \\
BND & A minimal, degenerate, or extreme object violates the map & Restrict the scope and handle the failed region separately \\
RT & The round trip loses proof-critical structure or merges distinct source objects & Change the encoding, add identifiers, or revise the source return \\
ACT & A matched-budget probe produces no new operation & Use a different target mechanism \\
RET & The target evidence does not cover the fixed source statement & Add a coverage lemma or change the source-side outcome \\
ENV & Dependency, version, resource, or cache failure & Repair the environment and replay the same artifact \\
BUD & The budget ends without a new admissible piece of evidence & Add budget, external evidence, or an updated bridge library \\
\bottomrule
\end{tabularx}
\label{tab:failure-codes}
\end{table}

When a route stops, its remaining budget returns to the controller for the same source task. The main experiment disables transfers across tasks because they would change task-level budgets. A reopening event must cite the earlier failure record and identify the proof-critical field that changed. Environment repairs retain the route version, while changes to the mathematical map, scope, or source return create a new version. A route stops after two consecutive stages without new independently checkable evidence, even if its agents can continue producing explanatory prose.

\section{Reproducibility bundle specification}
\label{app:release-package}

The reproducibility bundle is organized first by source task and then by run. A task directory contains the fixed statement, source, contamination cutoff, and license metadata. A run directory contains its configuration, route lineage, stress-test records, mathematical artifacts, verification records, source-side decision, cost data, and event log. A manifest maps every aggregate table cell to the corresponding set of per-task records.

\begin{table}[H]
\centering
\caption{Directories and review entry points in the reproducibility bundle}
\small
\begin{tabularx}{\linewidth}{p{3.1cm}p{4.2cm}X}
\toprule
Directory & Main files & Reproducible check \\
\midrule
\texttt{contracts/} & Fixed statement, exact negation, boundaries, and evidence contract & Check the denominator and statement version \\
\texttt{routes/} & Bridge records, route lineage, retrieval sources, and failure ledger & Reconstruct the candidate pool and route identities \\
\texttt{stress\_tests/} & Six stress-test records, object witnesses, and costs & Recompute the survival funnel and stopping decisions \\
\texttt{artifacts/} & Proofs, counterexamples, programs, Lean files, and output summaries & Check the mathematical objects independently \\
\texttt{verification/} & Target, bridge, replay, semantic, and source-side verification records & Check evidence scope and composition \\
\texttt{outcomes/} & Six outcome categories, blinded decisions, and release tiers & Reconstruct primary and error endpoints \\
\texttt{costs/} & Model, tool, wall-clock, and expert-time records & Recompute combined costs and sensitivity analyses \\
\texttt{events/} & Proposal, revision, withdrawal, recovery, and release events & Replay the final claim graph from an empty state \\
\bottomrule
\end{tabularx}
\label{tab:release-package}
\end{table}

The reproducibility bundle assigns 7 verified resolutions to the complete tier, which contains all layers listed above. Three witness-based resolutions omit license-restricted retrieval context but retain the statement, witness, source return, and independent reproduction procedure. Three delayed-release resolutions initially retain the evidence hash, evidence type, and review state, with the remaining artifacts scheduled after communication with the authors. All three tiers use the same outcome contract, and the main analysis can be reconstructed from their per-task records.

\section{Bridge-distance annotation}
\label{app:distance}

Two annotators score five dimensions as 0, 1, or 2 without seeing the target result. Total scores from 0 to 4 are adjacent-domain, scores from 6 to 10 are distant-domain, and a score of 5 enters a gray zone. Gray-zone routes participate in the full route pool and continuous trend analysis but are excluded from the primary binary comparison.

\begin{table}[H]
\centering
\caption{Five-dimensional anchors for bridge distance}
\small
\begin{tabularx}{\linewidth}{p{2.4cm}XXX}
\toprule
Dimension & Score 0 & Score 1 & Score 2 \\
\midrule
Object type & Same object and encoding & Standard derived object or specialization & Changed object type and combinatorial structure \\
Theory & Same theory & Standard adjacent branch & Different basic definitions and theorem system \\
Governing invariant & Original invariant retained & Classical associated quantity & New governing invariant \\
Correspondence length & One standard equivalence & Two or three established steps & New composition of maps required \\
Expert community & Complete review within one domain & Joint review by adjacent domains & Primary review by another community \\
\bottomrule
\end{tabularx}
\label{tab:distance-rubric}
\end{table}

Distance annotation does not use semantic embeddings, model confidence, target-side validation, or the final source outcome. Operational gain and certificate performance enter $A(B)$ and $K(B)$ separately.

\begin{table}[H]
\centering
\caption{Anchors for the six bridge-utility components on the 0 to 4 scale}
\small
\begin{tabularx}{\linewidth}{p{2.2cm}XXX}
\toprule
Component & Score 0 & Score 2 & Score 4 \\
\midrule
$A$ & No new operation & New verification procedure or local algorithm & Decisive algorithm, strong invariant, or proof interface \\
$K$ & No reduction in evidence length & Some obligations are compressed & Target certificate plus return is substantially shorter \\
$N$ & An equivalent tool exists in the source routes & The source tools can simulate it only indirectly & Neither the direct nor adjacent-domain pool has a comparable operation \\
$L$ & All proof-critical structure is preserved & Exceptions or merged objects can be enumerated & The map loses structure required by the source conclusion \\
$C_R$ & One standard source-return step & Several local obligations are needed & Source return nearly repeats the original proof \\
$C_D$ & A small example decides the route & A local program or expert check is required & Cost approaches that of a complete target proof \\
\bottomrule
\end{tabularx}
\label{tab:utility-anchors}
\end{table}

For the main analysis, every raw component score is divided by 4 before entering $U(B)$. Thus all six components lie in $[0,1]$. The coefficients are fixed to $\alpha=\beta=\gamma=\lambda=\mu=\nu=1$, with $A$, $K$, and $N$ entering positively and $L$, $C_R$, and $C_D$ entering negatively as specified in Section~\ref{sec:method}.

\section{Stress-test record template}
\label{app:pressure}

A stress-test record contains the bridge identifier and version, test type, input object, expected preservation relation, test method, result code, concrete witness, scope, cost, new obligations, and reopening conditions. The result code is \texttt{PASS}, \texttt{FAIL}, or \texttt{UNKNOWN}. A \texttt{PASS} records the covered scope, a \texttt{FAIL} includes an object or derivation, and an \texttt{UNKNOWN} identifies the smallest next task.

\begin{longtable}{p{2.1cm}p{3.23cm}p{3.23cm}p{3.23cm}}
\caption{Operational template for bridge-specific stress tests}\label{tab:pressure-template}\\
\toprule
Type & Fixed input & Preferred test & Output object \\
\midrule
\endfirsthead
\toprule
Type & Fixed input & Preferred test & Output object \\
\midrule
\endhead
Direction & $s,t$ and required result type & Implication direction and contrapositive & Valid result direction or reverse task \\
Assumption & $\Gamma_s$, target premises, and $\Delta^+$ & Premise difference and missing-assumption witness & New subproblem or conflicting object \\
Boundary & Minimal, degenerate, and extreme objects & Low-order enumeration and symbolic boundaries & Boundary certificate or counterexample \\
Round trip & $f,g$ and preserved quantities & Compare $g(f(x))$ and invariants & Preservation proof or information loss \\
Operation & $\mathcal A_t$ and source-tool inventory & Matched-budget paired probe & New operation and incremental artifact \\
Source return & Fixed source-return template & Obligation filling and coverage check & Source-side chain or open branch \\
\bottomrule
\end{longtable}

The same template applies to formalization bridges from natural language to Lean. The formal statement, quantifier map, permitted axioms, compilation environment, and semantic correspondence enter the same fields. The Lean kernel checks only the target operation.

\section{Capability matrix for experimental conditions}
\label{app:capability}

\begin{table}[H]
\centering
\caption{Capabilities and source-return rules in the route-pool conditions}
\scriptsize
\begin{tabularx}{\linewidth}{p{2.5cm}ccccccX}
\toprule
Condition & Direct & Adjacent pool & Distant pool & Retrieval & Target operation & Specific tests & Source-return rule \\
\midrule
Direct search & 1&0&0&0&0&0&Direct source-side review \\
Adjacent-domain bridges & 1&1&0&0&1&0&General source return \\
Distant-domain random & 1&0&1&0&1&0&General source return \\
Distant-domain retrieval & 1&0&1&1&1&0&General source return \\
Combined, no specific tests & 1&1&1&1&1&0&General source return \\
Full system & 1&1&1&1&1&1&Bridge-specific source return \\
\bottomrule
\end{tabularx}
\label{tab:capability}
\end{table}

The six conditions have identical per-task limits on total tokens, tool slots, target documents, cost, and expert minutes. When a route is rejected early, its released budget can move only within the same task. Distant-domain random selection and distant-domain retrieval share the target-mechanism catalog and minimum eligibility screen.

\begin{table}[H]
\centering
\caption{Paired outcomes for the full system and distant-domain retrieval}
\small
\begin{tabular}{lrrr}
\toprule
 & Distant retrieval resolved & Distant retrieval unresolved & Total \\
\midrule
Full system resolved & 7 & 6 & 13 \\
Full system unresolved & 1 & 106 & 107 \\
Total & 8 & 112 & 120 \\
\bottomrule
\end{tabular}
\label{tab:paired-main}
\end{table}

\section{Dataset construction protocol}
\label{app:harvest}

Let $T_h$ denote the frozen harvest cutoff. The author-eligibility window is the closed 24-month interval ending at $T_h$; it does not move with the publication or reading date of this paper. JCTA is used as an external author frame because its stated scope spans finite and discrete structures across several combinatorial subfields and its editorial policy requires a research-level contribution \citep{jcta_scope}. The frame is designed to target active researchers at a recognized specialist level, not to rank individuals or define all of combinatorics.

Dataset construction has seven steps.

\begin{enumerate}[leftmargin=1.8em,itemsep=4pt]
  \item Enumerate JCTA research articles whose first public publication date lies in the 24-month window; exclude editorials, corrigenda, and other non-research items.
  \item Build the author frame from those articles and disambiguate identities using names, affiliations, ORCID records when available, and publication histories.
  \item Collect those authors' publicly available papers from the same window, recording the first public date and version.
  \item Extract explicit conjectures, questions, and open problems, together with their contextual definitions and quantifiers.
  \item Deduplicate by mathematical contract while retaining source chains and wording variants.
  \item Screen eligibility without access to model attempts, then sample 30 tasks from each of the four structural strata.
  \item Fix the statement, retrieval cutoff, evidence types, budget, boundaries, and source-side decision criteria.
\end{enumerate}

The specification records an inclusion or exclusion reason for every candidate. Common exclusions are an unrecoverable definition, a solution before the cutoff date, an open research direction without a decidable statement, source licensing that prevents retention of the required text, or the inability to construct any eligible direct route during the evaluation period. Execution failures remain in the eligible denominator and do not trigger replacement.

\subsection{Contamination audit and replacement rule}

The audit distinguishes an overlap alert from substantive contamination. Exact statement overlap, a close paraphrase, a title match, or a shared named object raises an alert. Two curators then determine, without seeing system outcomes, whether the material contains a proof idea, a counterexample, an answer-bearing theorem, or a query cue that would materially shorten the route. Only this second category triggers replacement or blocking. The decision is frozen before execution, and each affected root task retains a link to its replacement.

\begin{table}[H]
\centering
\caption{Contamination decisions made before the evaluation set was frozen}
\small
\begin{tabularx}{\linewidth}{p{3.1cm}rX}
\toprule
Audit stage & Count & Decision and recorded evidence \\
\midrule
Overlap alert & 146 & Retain the matched text, date, version, query, and alert type for curator review \\
Substantive answer cue & 23 & Record the proof idea, counterexample clue, answer-bearing theorem, or query shortcut \\
Root-task contamination & 18 & Remove the task before execution and link it to a same-stratum replacement \\
Route-level contamination & 5 & Keep the root task but block the affected route and its answer-bearing material \\
Restoration check & 23 & Reinsert only for sensitivity analysis; the full-system mean rises by 2.1 verified resolutions \\
\bottomrule
\end{tabularx}
\label{tab:contamination-audit}
\end{table}

Long-standing conjectures such as Goldbach and Hadwiger illustrate the motivation for this rule: their statements and surrounding ideas have been discussed publicly for decades, so success alone cannot separate fresh search from prior exposure when training corpora are not inspectable. The protocol therefore uses recency and date truncation to reduce exposure risk, while the restoration check estimates the direction and size of detected contamination in this dataset. It does not certify the absence of undetected contamination.

\subsection{Refutations and author communication}

Counterexamples to statements by living authors undergo three stages of checking. First, the generating route provides a concrete object and checks every premise. Second, an independent program or mathematician recomputes the result from the fixed statement without reading the generator's correctness assessment. Third, the source-side review checks the statement version, quantifiers, scope, and violated conclusion. Once all three agree, the evidence package contains the original statement, concrete witness, premise-by-premise check, and reproducible verification procedure.

The release order follows evidence maturity. A complete artifact bundle can be checked directly. If interpretation of the statement still requires confirmation from the author, the statement, witness, and reproduction procedure are first sent privately with a defined response window. If the statement version remains ambiguous, the outcome stays conditional or local. Communication is used only to resolve scope and wording.

\section{Verification and formalization}
\label{app:verification}

\subsection{Three-stage Lean procedure}

The premise-retrieval stage records its queries, Mathlib version, and candidate declarations. The candidate-generation stage records the model version, input, Lean file, and error feedback. The independent-replay stage compiles the candidate in the registered environment and records the declaration, imports, exit status, axiom list, and scope. The three stages have separate permissions, and the generator cannot update the replay result.

\begin{table}[H]
\centering
\caption{Required fields in a Lean verification record}
\small
\begin{tabularx}{\linewidth}{p{3.0cm}X}
\toprule
Field & Content \\
\midrule
Declaration identity & Claim identifier, statement version, Lean name, and source file \\
Environment & Lean, Mathlib, project dependencies, execution policy, and imports \\
Result & Compilation status, error category, runtime, and resource limit \\
Axiom audit & \texttt{sorry}/\texttt{admit}, new axioms, \texttt{unsafe}, and \texttt{\#print axioms} \\
Semantic edge & Correspondence of natural-language objects, quantifiers, scope, degenerate cases, and direction \\
Composition edge & Parent claim, dependency subgraph, open obligations, and covered range \\
\bottomrule
\end{tabularx}
\label{tab:lean-receipt}
\end{table}

Environment errors have separate codes for unavailable dependencies, insufficient memory, corrupted caches, and version conflicts. Mathematical feedback distinguishes missing assumptions, type conflicts, definitional mismatch, failed rewriting, termination gaps, construction gaps, and counterexample witnesses. The same candidate may be replayed after an environment repair, while a mathematical revision creates a new version.

\subsection{Certificates from deterministic programs}

Finite counterexamples, enumerations, integer computations, and graph searches retain their inputs, outputs, code version, run configuration, and independent reproduction record. Final claims preferably cite compact certificates such as counterexample coordinates, subset lists, SAT proofs, factorizations, or coverage tables. Repeating the same implementation establishes repeatability but not independence. A second path uses another implementation, a deterministic verification procedure, a manual check, or formalization of the critical step.

The program scope is matched to the scope of the statement. An enumeration for $n\leq20$ supports a finite-range result. If it contributes to a general proof, the composition obligations identify the theoretical step that continues beyond that range. Randomized programs may explore candidates, but a root conclusion cites either a reproducible concrete witness or a fixed statistical protocol.

\section{Persistent-state experiment}
\label{app:persistence}

\begin{table}[H]
\centering
\caption{State experiment after three recoveries and two version changes}
\small
\resizebox{\linewidth}{!}{%
\begin{tabular}{lrrrr}
\toprule
State condition & Repeated target runs & Stale returns & Recovery minutes & Unrelated withdrawals \\
\midrule
No persistent state & 29 & 7 & 93 & 25 \\
Task graph only & 13 & 4 & 56 & 14 \\
Task and claim graphs & \textbf{4} & \textbf{0} & \textbf{34} & \textbf{5} \\
\bottomrule
\end{tabular}}
\label{tab:persistence-app}
\end{table}

The first recovery restarts only the execution process. Before the second recovery, one bridge assumption is changed; before the third, one target theorem is upgraded. A preregistered reference dependency table records the claims that should be affected, and the system output is compared with that reference. Unrelated withdrawals count claims that do not depend on the changed object but are nevertheless rechecked.

\section{Mathematical certificates for two case studies}
\label{app:case-certificates}

\subsection{Zhao counterexample certificate}
\label{app:zhao-certificate}

For a finite abelian group $H$, let $s_{\leq k}(H)$ be the least integer $\ell$ such that every sequence over $H$ of length at least $\ell$ has a nonempty zero-sum subsequence of length at most $k$. Zhao's Conjecture~6.1 states that, if $H$ has rank at least two, $D(H)=D^*(H)$, $H\neq C_2^4$, $D(H)-2\geq\exp(H)$, and $\exp(H)<(D(H)-1)/2$, then \citep{zhao2025zerosum}
\[
s_{\leq D(H)-2}(H)=D(H)+1.
\]

Write $G=C_2\oplus C_4^3$ in coordinates $(u;x,y,z)$ modulo $(2,4,4,4)$. Olson's formula for finite abelian $p$-groups gives
\[
D(G)=D^*(G)=1+(2-1)+3(4-1)=11.
\]
The group has rank four, is not $C_2^4$, has exponent $4$, and satisfies $D(G)-2=9\geq4$ and $4<(D(G)-1)/2=5$. Thus it satisfies every hypothesis of the conjecture.

Define
\[
a=(0;1,0,0),\quad b=(0;0,1,0),\quad c=(0;0,0,1),\quad T=(1;3,3,3),
\]
put $p=T-a$, $q=T-b$, and $r=T-c$, and let $S=a^3b^3c^3pqr$. For a selected subsequence, let $i,j,k\in\{0,1,2,3\}$ be the multiplicities of $a,b,c$ and let $\alpha,\beta,\gamma\in\{0,1\}$ indicate whether $p,q,r$ are selected. If $Q=\alpha+\beta+\gamma$, the sum is
\[
(i-\alpha)a+(j-\beta)b+(k-\gamma)c+QT.
\]
Consider
\[
\phi:C_4^4\to G,\qquad
\phi(x,y,z,t)=(t\bmod2;x+3t,y+3t,z+3t\bmod4).
\]
Its kernel is $\{(0,0,0,0),(2,2,2,2)\}$. The zero vector forces the empty selection. The nonzero vector forces
\[
Q=2,\qquad i=2+\alpha,\qquad j=2+\beta,\qquad k=2+\gamma,
\]
and hence a selected length of $i+j+k+Q=10$. Conversely, each selector satisfying these equations has coefficient vector $(2,2,2,2)$ and is zero-sum. There are nine indexed zero-sum subsequences: choose which of $p,q,r$ is omitted, then choose two of the three copies in the corresponding repeated block. Every nonempty zero-sum subsequence of $S$ therefore has length $10$. The length-12 sequence $S$ has none of length at most $9$, so
\[
s_{\leq9}(G)\geq13>12=D(G)+1.
\]
This concrete inequality is the source-side contradiction. The exhaustive $4095$-subset computation independently checks the finite classification above.

\subsection{AJT(5) sparse-row theorem}
\label{app:ajt-certificate}

Let
\[
T_m=(\mathbb F_5^*)^m,\qquad
N_m(B,b)=\#\{x\in T_m:Bx+b\in T_m\},
\]
and define $a_1=3$, $a_2=8$, and $a_m=8a_{m-2}$ for $m\geq3$. If $B\in\mathrm{GL}_m(\mathbb F_5)$ and every row of $B$ has support at most two, then
\[
N_m(B,b)\geq a_m
\]
for every $b\in\mathbb F_5^m$.

Associate a row-support multigraph to $B$: variables are vertices, rows with two nonzero entries are edges, and rows with one nonzero entry are loops. Row and column permutations separate the connected components. If a component has $r$ rows and $v$ vertices, its block has rank at most both $r$ and $v$. Since the total rank, row count, and vertex count are all $m$, every component block satisfies $r=v=\operatorname{rank}(B_i)$. Each connected component is therefore unicyclic, with loops and parallel-edge 2-cycles allowed.

Remove all leaves that do not lie on a cycle. Reinstating a leaf requires an affine inequality $\alpha x+d\neq0$ with $\alpha\neq0$ and $x\in\mathbb F_5^*$, which leaves at least three choices for any $d$. For an oriented cycle of length $\ell$, write the forbidden relations as $x_{i+1}=g_ix_i+c_i$. On the four nonzero field elements, each relation is a partial permutation $Q_i$ that extends to a permutation matrix $P_i$. If $J$ is the all-ones matrix, then $J-Q_i\geq J-P_i$ entrywise. Nonnegative matrix products and traces preserve this inequality, giving
\[
\operatorname{tr}\!\prod_i(J-Q_i)
\geq
\operatorname{tr}\!\prod_i(J-P_i)
=3^\ell+(-1)^\ell(f-1),
\]
where $0\leq f\leq4$ is the number of fixed points of $\prod_iP_i$. For every $\ell\geq2$, this is at least $8\cdot3^{\ell-2}$; a loop component contributes at least $3^{n_i}$.

Different blocks use disjoint variables, so their counts multiply. If $c$ is the number of non-loop cycle components, then
\[
N_m(B,b)\geq8^c3^{m-2c}
\geq8^{\lfloor m/2\rfloor}3^{m-2\lfloor m/2\rfloor}=a_m,
\]
because $c\leq\lfloor m/2\rfloor$ and the expression decreases with $c$. The proof covers every affine offset $b$ and every dimension. AJT(5) is the special case $B=A$ and $b=0$ \citep{alon1989nowherezero}. Dense rows are outside the theorem's scope, so the general AJT(5) problem remains open.

\section{Model identities and matched costs}
\label{app:models}

Models A, B, and C in the main text correspond to GPT-5.6, Claude 4.8, and DeepSeek. Each matched-cost condition has the same per-task limits on input tokens, output tokens, tool slots, retrieved documents, and total cost. A single-model condition may fill several roles, but each role uses an isolated context. The homogeneous multi-agent condition adds role instances while reducing their individual budgets to preserve the total. The heterogeneous condition uses the same total budget and permits verification records to pass between models.

Cross-model handoffs are counted by bridge identifier. Replacing a model without changing the object map, target mechanism, or source-return direction leaves the bridge identity unchanged. A verified resolution counts as a cross-model handoff only when at least two of bridge proposal, decisive revision, target validation, and source-side review are performed by different models.

\section{Cost accounting}
\label{app:cost}

Each task records model input, output, caching, and retries; retrieval, Lean, deterministic programs, computer algebra, SAT/SMT, and domain-specific verification; local CPU or GPU time and wall-clock time; minutes spent on bridge construction, target-domain review, source-side review, state maintenance, and environment maintenance; and one-time development time for task curation, the bridge library, retrieval, stress-test generation, and the execution system.

The per-task combined cost is
\[
C_{\mathrm{joint}}=C_{\mathrm{machine}}+
r_h\frac{T_{\mathrm{bridge}}+T_{\mathrm{review}}+T_{\mathrm{state}}}{60}.
\]
The main analysis sets $r_h=200$ dollars per hour and repeats the calculation at 100 and 400 dollars. For each experimental condition, the median-normalized cost per verified resolution is
\[
C_{\mathrm{resolution}}=
\frac{\operatorname{median}_{i}(C_{\mathrm{joint},i})}
{\widehat p_{\mathrm{resolution}}},
\]
where $\widehat p_{\mathrm{resolution}}$ is the verified-resolution rate for that condition. Failed runs, unresolved tasks, and incorrect conclusions remain in the per-task cost distribution.

One-time development required 552 person-hours: 146 for integrated execution and data contracts, 118 for the bridge catalog and retrieval, 104 for stress-test generation and source return, 96 for verification services, and 88 for analysis and release procedures. The bridge-reuse analysis changes only the construction minutes for a new task and does not retroactively amortize development time over completed tasks.

\section{Reproduction protocol}
\label{app:reproduction}

The reproducibility bundle is organized by run identifier. Each run directory contains its configuration, task version, route pool, stress-test records, target artifacts, verification records, source-side outcome, blinded decision, and cost row. The main tables and quantitative figures are generated from the same per-task records. Reproduction follows these steps:

\begin{enumerate}[leftmargin=1.8em,itemsep=4pt]
  \item Load the 120 fixed tasks and the target-mechanism catalog.
  \item Reconstruct the direct, adjacent-domain, distant-domain random, and distant-domain retrieval pools with the fixed seeds.
  \item Run eligibility screening, the six stress tests, and budget reallocation.
  \item Replay target-side verification, Lean artifacts, and deterministic programs in isolated environments.
  \item Obtain blinded source-side decisions, then run the paired tests and interval estimates.
  \item Rebuild the main tables, figures, failure ledger, and cost analysis from the per-task outcomes.
\end{enumerate}

Software metadata records the official version name or release tag, source URL, verification date, run identifier, configuration name, and decisive configuration fields. Failures, timeouts, invalid operations, and human stops retain their original denominator and result code.

\clearpage
\section{Full Gao candidate-proof manuscript}
\label{app:gao-full}

The following 13 pages reproduce, without textual revision, the manuscript
\emph{The Gao Constant of Generalized Dihedral Groups with an Abelian
$p$-Group Kernel, $p$ Odd}. The snapshot is the PDF delivered by GitHub
PR~\#7, \emph{Rewrite Gao manuscript as a paper-first auditable arXiv
source}, from branch \texttt{paper/arxiv-rewrite-2026-08-24} at commit
\texttt{6d4ab81}. The pull request and its review state are available at
\url{https://github.com/randomcat4/gao0824/pull/7}. The complete companion
LaTeX source and build records are included under
\texttt{supplement/gao\_pr7/source/} in this submission package.

This reproduction is an audit surface, not a correctness certificate. The
companion repository classifies the text as a complete natural-language
candidate proof pending independent line-by-line review, verification of its
source-derived theorem instances, and completion of the top-level Lean
obligation. Statements inside the reproduced manuscript such as ``we
determine'' belong to that candidate manuscript and do not change the
source-side outcome reported in this paper.

\includepdf[
  pages=-,
  scale=0.96,
  pagecommand={\thispagestyle{empty}}
]{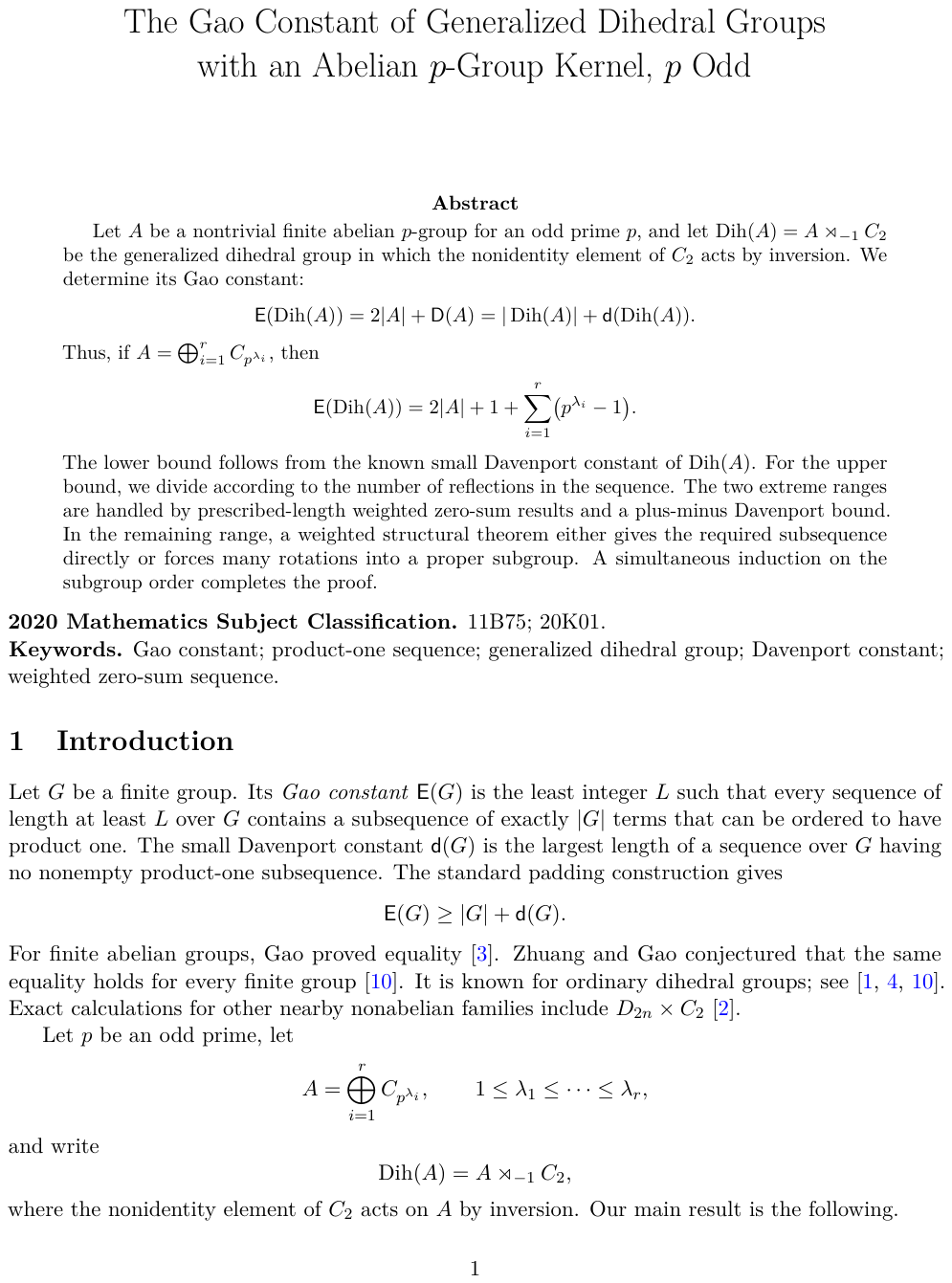}

\end{document}